%% file: main.tex
\documentclass{article}

\usepackage{microtype}
\usepackage{graphicx}
\usepackage{booktabs} 

\usepackage{url}

\usepackage[utf8]{inputenc} 
\usepackage[T1]{fontenc}    
\usepackage{booktabs}       
\usepackage{amsfonts, amsmath, amssymb}       
\usepackage{nicefrac}       
\usepackage{microtype}      
\usepackage{xcolor}         
\usepackage{wrapfig}
\usepackage{subfig}
\usepackage{float}
\usepackage{xspace}
\usepackage{color,soul}
\usepackage{multirow}
\usepackage{subfig}
\usepackage{tikz}
\usepackage{circledsteps}
\usepackage{formatting/macros}
\usepackage{formatting/results}
\usepackage{enumitem}
\usepackage{cancel}

\usepackage{hyperref}

\usepackage[accepted]{mlsys2024}

\mlsystitlerunning{Keyformer: KV Cache reduction through key tokens selection for Efficient Generative Inference}

\begin{document}

\twocolumn[
\mlsystitle{Keyformer: KV Cache reduction through key tokens selection for Efficient Generative Inference}



\mlsyssetsymbol{equal}{*}

\begin{mlsysauthorlist}
\mlsysauthor{Muhammad Adnan}{equal,to}
\mlsysauthor{Akhil Arunkumar}{goo}
\mlsysauthor{Gaurav Jain}{goo}\\
\mlsysauthor{Prashant J.~Nair}{to}
\mlsysauthor{Ilya Soloveychik}{goo}
\mlsysauthor{Purushotham Kamath}{goo}
\end{mlsysauthorlist}

\mlsysaffiliation{to}{Department of Electrical and Computer Engineering, The University of British Columbia, Vancouver, BC, Canada}
\mlsysaffiliation{goo}{d-Matrix, Santa Clara, California, USA}

\mlsyscorrespondingauthor{Muhammad Adnan}{adnan@ece.ubc.ca}

\mlsyskeywords{Machine Learning, MLSys}

\pagenumbering{arabic}
\setcounter{page}{1}

\vskip 0.3in

\input{body/abstract}
]
\printAffiliationsAndNotice{\mlsysEqualContribution}

\input{body/introduction}
\input{body/challenges}
\input{body/keyformer}
\input{body/evaluation}
\input{body/related_work}
\input{body/future_work}
\input{body/conclusion}

\bibliography{Keyformer/references}
\bibliographystyle{mlsys2024}

\clearpage
\appendix
\input{body/appendix}


\end{document}

%% file: body/abstract.tex
\begin{abstract}

Transformers have emerged as the underpinning architecture for Large Language Models (LLMs). In generative language models, the inference process involves two primary phases: prompt processing and token generation. Token generation, which constitutes the majority of the computational workload, primarily entails vector-matrix multiplications and interactions with the Key-Value ($\mathsf{KV}$) Cache. This phase is constrained by memory bandwidth due to the overhead of transferring weights and \kv values from the memory system to the computing units. This memory bottleneck becomes particularly pronounced in applications that require long-context and extensive text generation, both of which are increasingly crucial for LLMs.

This paper introduces ``\keyformer'', an innovative inference-time approach, to mitigate the challenges associated with \kv size. \keyformer leverages the observation that approximately 90\% of the attention weight in generative inference focuses on a specific subset of tokens, referred to as ``key'' tokens. \keyformer retains only the \key in the \kv by identifying these crucial tokens using a novel score function. This approach reduces both the \kv size and memory bandwidth usage without compromising model accuracy. We evaluate \keyformer's performance across three foundational models: GPT-J, Cerebras-GPT, and MPT, which employ various positional embedding algorithms. Our assessment uses a variety of tasks, with an emphasis on summarization and conversation tasks involving extended contexts. We show that \keyformer reduces inference latency by \latencyimprv and improves token generation throughput by \throughputimprv, while preserving the model's accuracy.

\end{abstract}

%% file: body/introduction.tex
\section{Introduction}
\label{sec:introduction}

Transformers have proven to be particularly successful in tasks such as language modeling~\cite{bart, brownlanguage, text-transformer}, image recognition~\cite{vit}, recommendations~\cite{bert4rec, transformers4rec, adrec,llm_survey}, and text generation with the advent of Large Language Models (LLMs). Unfortunately, LLM deployment presents critical \emph{inference latency and throughput} concerns. This is primarily attributed to the sequential autoregressive nature of generative inference, particularly when handling inputs with larger contexts. Despite advancements, modern LLMs face challenges in efficiently processing longer input sequences, as evidenced by recent studies~\cite{longbench, longchat2023, longlora, efficient_attn_summarization}. Unfortunately, the increased memory and compute requirements associated with longer sequences exacerbate LLM inference latency and reduce throughput. This paper proposes inference-time strategies for accuracy-preserving, low-latency, high-throughput LLM systems.

    LLMs employ Transformers and rely on the `attention mechanism' to understand the relationships between words within a given input sequence~\cite{attention}. As the attention mechanism scales quadratically with the size of the input sequence, it tends to present the largest latency overhead during inference~\cite{adaptive_attention, flashattention, performers}.  Additionally, due to the autoregressive nature of token generation in LLMs, there is a need to \emph{recompute} key and value vectors for all previous tokens. To mitigate this, LLMs utilize a storage structure known as a Key-Value Cache (\kv)~\cite{fairseq}. \kv retains previously computed key-value pairs, eliminating the need for costly re-computation of these vectors.

\begin{figure}
  \centering
  \includegraphics[width=1\columnwidth]{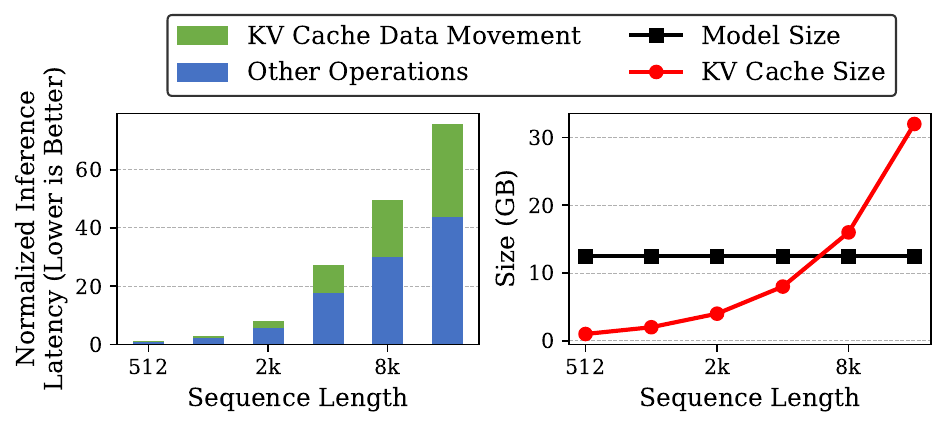}
  \vskip -0.15in
  \caption{(a) Inference latency normalized to sequence length of 512. We measure the \kv data movement for MPT-7B~\cite{mpt} model with varying sequence length (50\% context + 50\% text generation). (b) The \kv size and model size as sequence length varies. The studies are performed on an NVIDIA A100 GPU with a batch size of 1 and beam size of 4. 
}
\label{fig:motivation}
\vskip -0.25in
\end{figure}

However, \kv presents scalability challenges. Accessing the \kv from off-chip memory during token generation introduces additional memory latencies and is constrained by memory bandwidth limitations. For instance, in the MPT-7B model illustrated in Figure~\ref{fig:motivation}(a), increasing the sequence length by 16$\times$ (from 512 to 8K) results in a more than \emph{50$\times$ increase} in inference latency. Moreover, approximately 40\% of the total inference time (highlighted in green) is consumed by \kv data movement. Importantly, a larger context not only increases the size of the \kv but also prolongs the time required for other operations (depicted in blue). Similarly, as shown in Figure~\ref{fig:motivation}(b) for the MPT-7B model, the \kv size surpasses the model size when the sequence length exceeds 8K. Thus, \kv sizes present a roadblock to enabling low-latency, high-throughput inference for large sequences.

Previous studies have explored mitigating attention mechanisms' memory and computation requirements when dealing with longer sequences~\cite{bigbird, reformer, linformer, longformer}. While system-level optimizations like FlexGen~\cite{flexgen}, Flash Attention~\cite{flashattention}, Paged Attention~\cite{pagedattention}, and multi-dimensional partitioning~\cite{efficiently_scaling_trans_inf} aim to improve the scalability of generative AI, they often overlook the fundamental challenge of expanding \kv size. Techniques like multi-query~\cite{mqa} and group-query attention~\cite{gqa} propose reducing \kv size by eliminating specific attention heads from writing to the \kv, but these methods typically require \emph{resource-intensive model retraining or fine-tuning}. This becomes complex as various accelerators are already deployed in the field. Thus, there is a pressing need for inference-time techniques for \kv reduction. This is even more challenging as any proposed technique must conform to the strict constraints for model accuracy. For instance, \emph{MLPerf~\cite{mlperf} mandates that any optimization applied to LLMs maintain a model accuracy between 99\% to 99.9\% of the baseline}.

\begin{figure}
  \centering
  \includegraphics[width=1\columnwidth]{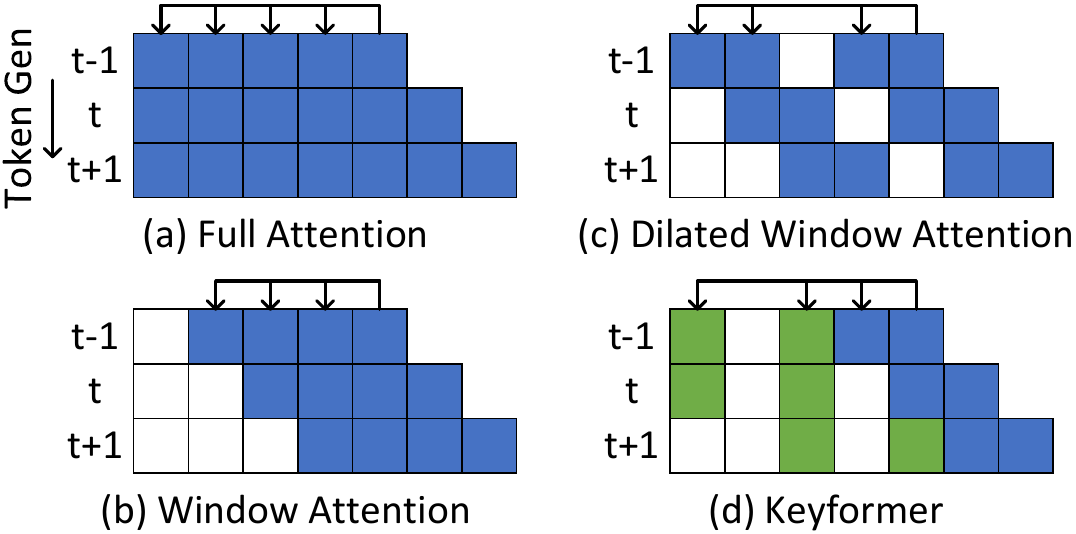}
  \vskip -0.1in
  \caption{Attention block for generative inference. (a) Full attention with current token attending all previous tokens. (b) Window attention ($w=4$): Focusing on the most recent 4 tokens. (c) Dilated window attention ($w = 4$, dilation = 1). (d) \keyformer ($w = 2$, $k = 2$): A mix of recent window (w) and \key (k). White color indicates no attention, while blue color indicates attention. The green color identifies the \key and their respective attention. The values of the three consecutive token generation iterations are $t-1, t, t+1$.
}
\label{fig:attention}
\vskip -0.3in
\end{figure}

\begin{figure*}[t]
  \centering
  \subfloat[Attention Sparsity]{
  \begin{minipage}[t]{0.333\textwidth}
  \includegraphics[width=1\textwidth]{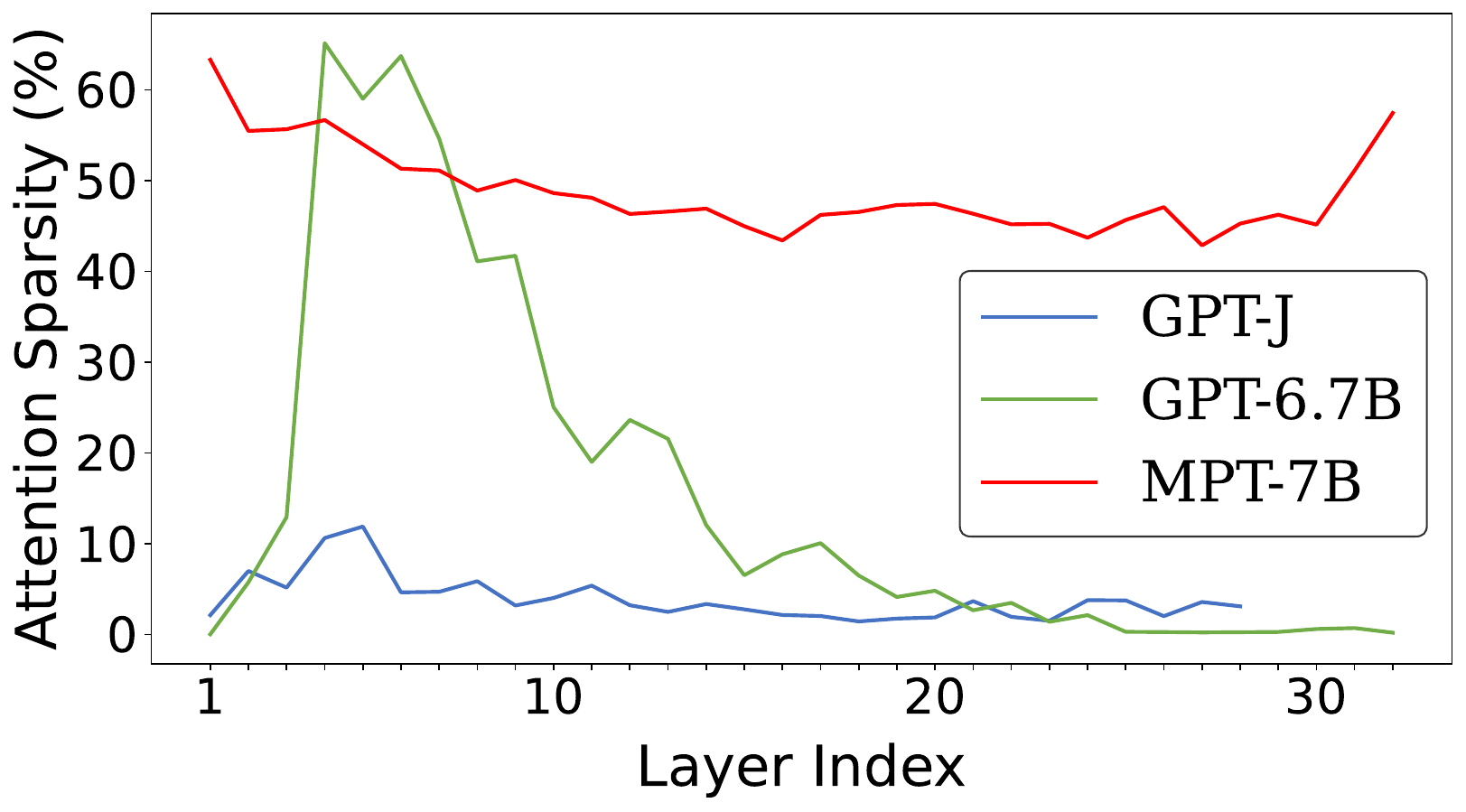}
  \vskip -0.05in
  \label{fig:sparsity}
  \end{minipage}
  }\hfill
  \subfloat[Average Attention Score]{
  \begin{minipage}[t]{0.3335\textwidth}
  \includegraphics[width=1\textwidth]{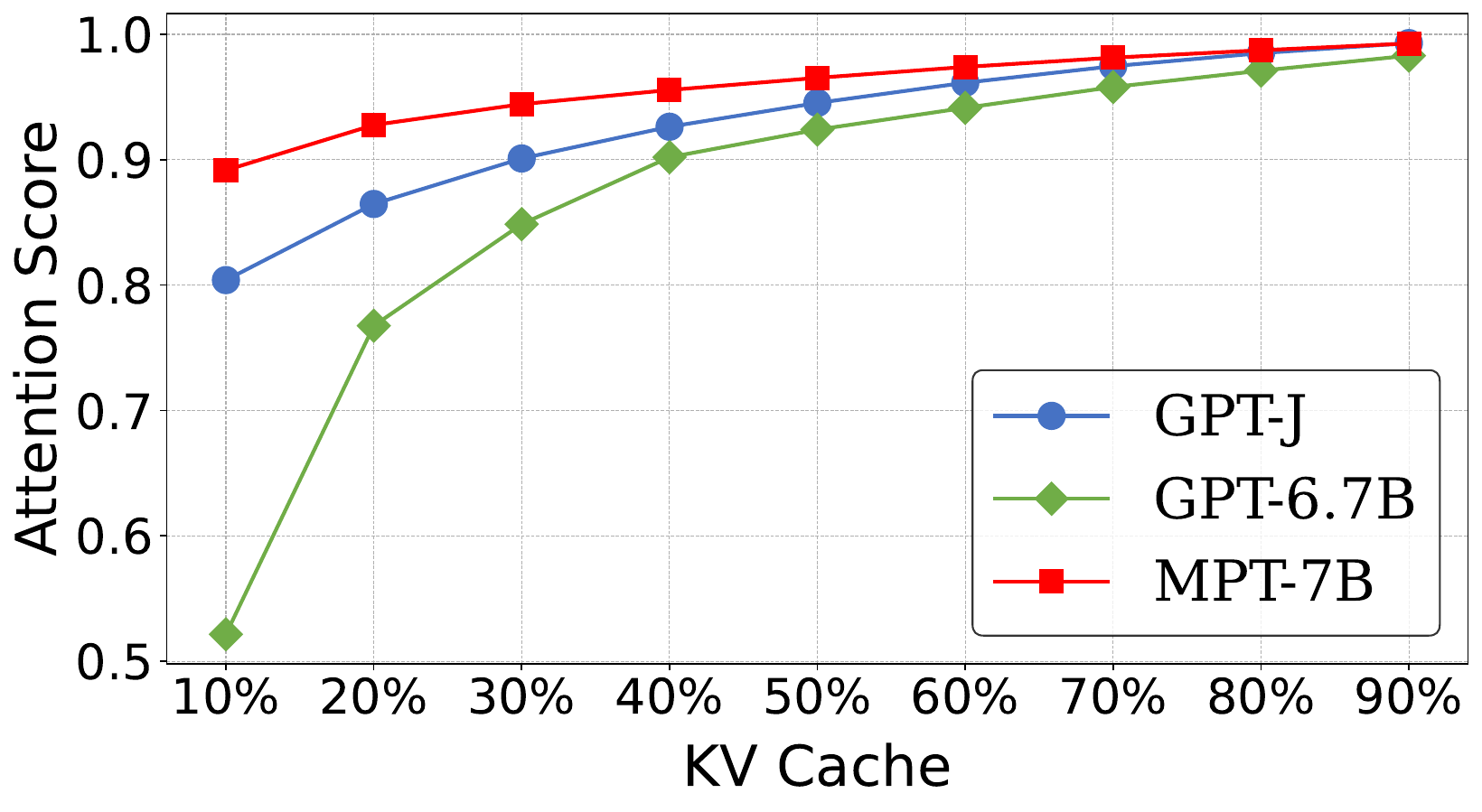}
  \vskip -0.05in
  \label{fig:attn_score}
  \end{minipage}
  }\hfill
  \subfloat[Accuracy]{
  \begin{minipage}[t]{0.29\textwidth}
  \includegraphics[width=1\textwidth]{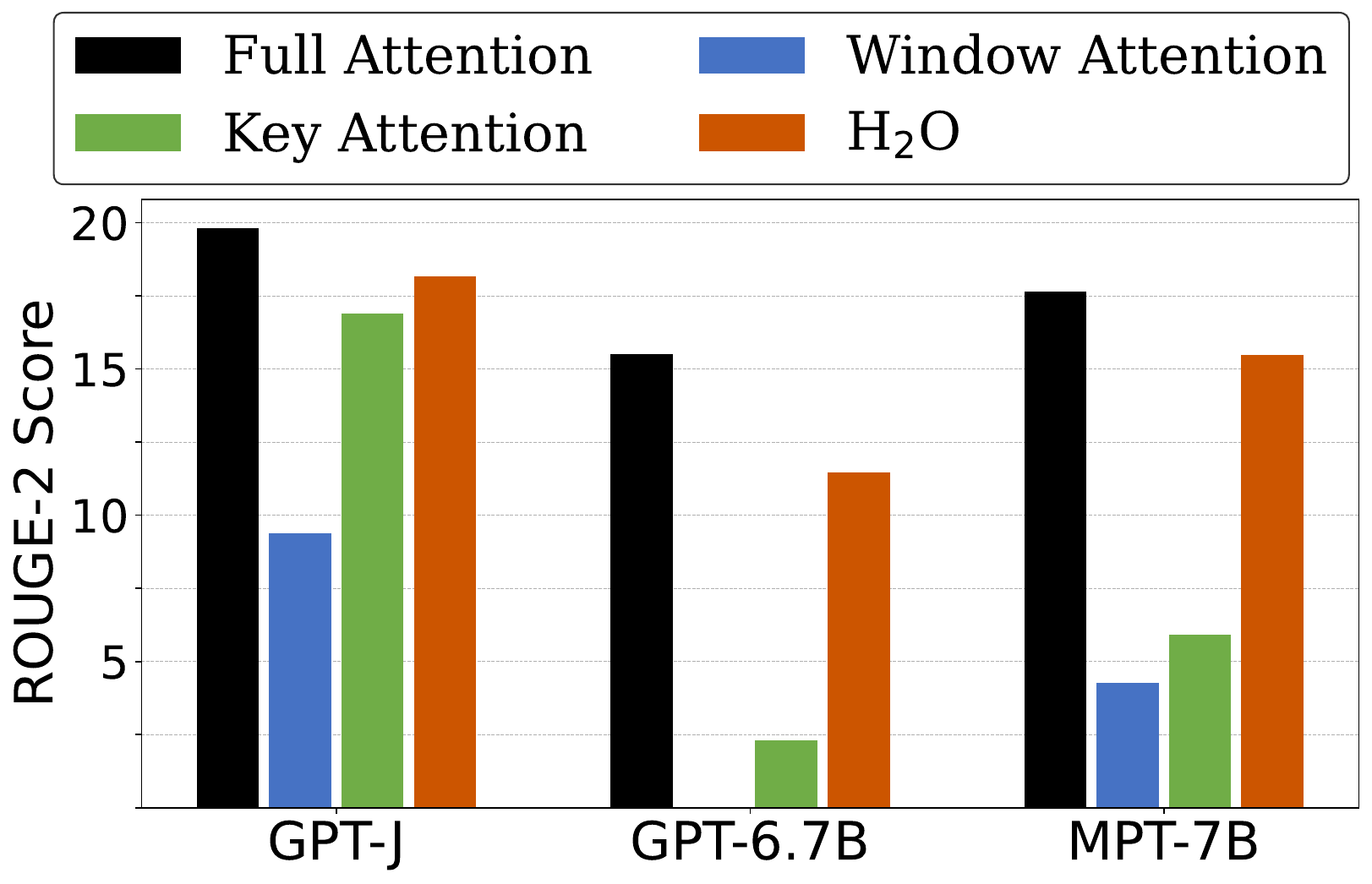}
  \vskip -0.05in
  \label{fig:attn_accuracy}
  \end{minipage}
  }\hfill
  \vskip -0.1in
  \caption{(a) Default attention sparsity across different models. (b) Average attention score of three different models with ~90\% of attention score dedicated to ~40\% of the tokens called \key. (c) Accuracy comparison of three models with different attention schemes. `Full Attention' uses the full \kv size, while `Window Attention' and `H$_{2}$O' use 50\% of the \kv size. All models use the CNN/DailyMail~\cite{cnn_dm} dataset for the summarization task.}
  \vskip -0.1in
\end{figure*}

To address these concerns, we introduce \keyformer~\footnote{\hyperlink{https://github.com/d-matrix-ai/keyformer-llm}{https://github.com/d-matrix-ai/keyformer-llm}.}, a novel method for dynamically reducing the \kv size during inference. \keyformer does this by intelligently discarding unnecessary tokens without losing accuracy. The critical insights of \keyformer are demonstrated in Figure~\ref{fig:attention}, where we also compare with existing state-of-the-art inference optimizations. Figure~\ref{fig:attention}(a) illustrates the traditional `Full Attention'~\cite{brownlanguage} mechanism, where each newly generated token attends to \emph{all} preceding tokens in the sequence. Figure~\ref{fig:attention}(b) depicts `Window Attention,'~ \cite{sparse_transformer} which maintains a fixed-size \emph{sliding window} of recent tokens, thereby reducing the size of the \kv. However, this method restricts the model's capacity to capture comprehensive semantic information from the past, leading to lower-quality text generation and decreased accuracy. Figure~\ref{fig:attention}(c) presents a variant called `Dilated Window Attention,' with similar accuracy limitations to windowed attention.

To address this, \keyformer leverages the insight that certain tokens carry more significance than others. Specifically, it observes that \emph{nearly 90\% of the attention weight focuses on a small subset known as \key}. These tokens are crucial for LLMs to grasp context but may fall outside the sliding window of window attention. \keyformer introduces a mixed attention approach, depicted in Figure~\ref{fig:attention}(d), which combines recent tokens with the preceding \key when generating the next token. Our experiments show that \keyformer demonstrates significant improvements over state-of-the-art methods such as H$_{2}$O~\cite{h2o}. This is because, unlike H$_{2}$O, which identifies ``heavy hitters'' solely based on attention scores, \keyformer considers the importance of discarded tokens in identifying \key.

We evaluate \keyformer on multiple models, including GPT-J~\cite{gptj}, Cerebras-GPT~\cite{cerebrasgpt}, and MPT~\cite{mpt}, across various tasks like summarization and conversation for long sequences. Even with a 50\% reduction in \kv, \keyformer preserves accuracy while reducing inference latency by \latencyimprv and boosting token generation throughput by \throughputimprv.

%% file: body/challenges.tex
\section{Background and Motivation}
\label{sec:problem}

\subsection{Inference Process in Large Language Models}
In language modeling, the task involves estimating the probability of the next token based on preceding tokens $x_{1}, x_{2}, \ldots, x_{n}$. For generative Large Language Models (LLMs), the inference process unfolds in two phases:
\setlist{nolistsep}
\begin{enumerate}[leftmargin=*,noitemsep]
    \item \textbf{Prompt Processing Phase:} This phase helps input context undergo causal processing, enabling the model to generate keys and values for all tokens within the context. These key-value pairs are then stored in the \kv.
    \item \textbf{Token Generation Phase:} This phase sequentially and auto-regressively generates text. Each token is produced by passing through all layers of the generative model. Notably, the generation of the next token relies on the previously generated tokens and their order.
\end{enumerate}

To enhance inference efficiency, repeated and complex computations of Key ($\mathsf{K}$) and Value ($\mathsf{V}$) tensors across all layers are avoided by caching these tensors. This is referred to as the \kv. The \kv is sequentially populated during each token generation step~\cite{strati2024dejavu}, until the text generation process is completed.

\subsection{Reducing KV Cache Size by Exploiting Sparsity}

To address the challenge posed by the expanding \kv, let us examine a sequence, denoted as $S_{n}$, comprising $n$ tokens and its \kv contents for a single attention head and layer. In full attention, the \kv components involve $n$ keys and values. These grow proportionally with $S_{n}$. To mitigate this, we can shrink the \kv size to accommodate shorter sequences, designated as $S_{k}$. This involves using a reduced number of tokens, transitioning from $n$ to $k$, where $S_{k}$ is a subset of $S_{n}$, and $k$ is less than $n$. This reduction can be achieved by leveraging the inherent sparsity within the attention mechanism of LLMs.

Despite the substantial computational demands during the training of transformers, there exists inherent sparsity within the attention mechanism. However, the extent of sparsity may vary depending on the particular downstream task. Figure~\ref{fig:sparsity} illustrates the diverse levels of attention sparsity among different models utilized for summarization tasks with the CNN/DailyMail dataset. This variability manifests across various levels of the model, including the overall model, individual layers, and distinct sections of the model.

\subsection{Improving Performance by Using Key Tokens}

In Figure~\ref{fig:attn_score}, the Cumulative Distribution Function (CDF) depicts the relationship between attention score and the fraction of the total context. Notably, a small subset of tokens receives the most attention during text generation. This underscores the significance of specific \key and their pivotal role in comprehending context and facilitating text generation. However, dynamically determining which tokens serve as \key, especially in cases where the input sequence contains unknown or unseen tokens during inference, presents a considerable challenge.

\subsubsection{Leveraging Score Function to Identify Key Tokens}

We introduce a score function $f_{\theta}$ for each token to identify the $k$ \key out of a total of $n$ tokens. In the multi-head attention mechanism, attention scores determine the degree of connection between a single token and all other tokens. This is described by Equation~\ref{eqn:1}.
\begin{equation}\label{eqn:1}
\text{Attention\;Score} = \mathrm{softmax} \left( \frac{QK^T}{\sqrt{d_k}} \right)
\end{equation}

\begin{figure}[ht!]
  \centering
  \includegraphics[width=1\columnwidth]{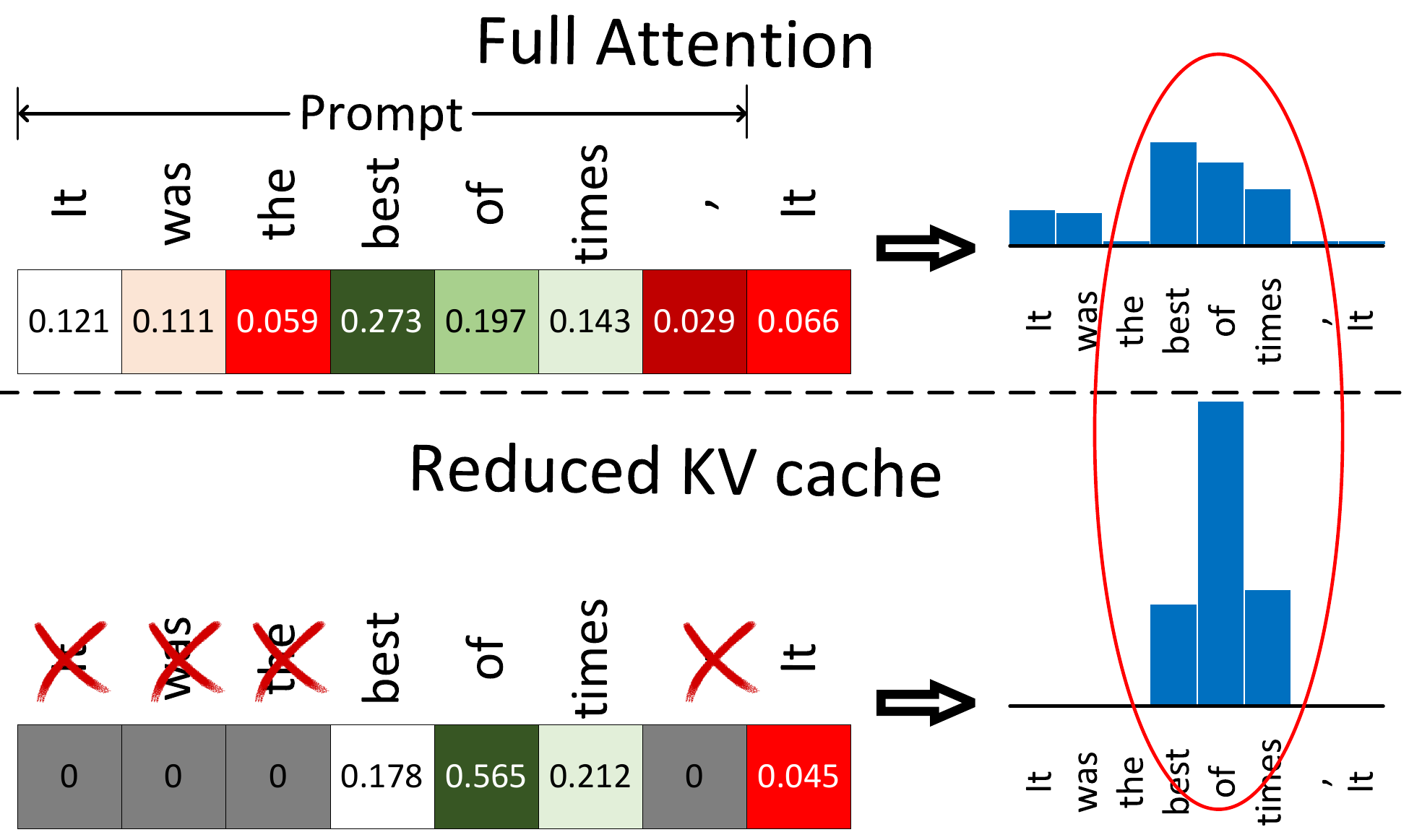}
  \caption{Reducing the \kv introduces a change in the distribution of attention scores. As tokens are removed, their distribution becomes uneven among the remaining cached tokens. Thus, it affects the identification of \key according to the score function $f_{\theta}(acc\;attn)$. The figure shows this effect for the attention scores for the MPT-7B~\cite{mpt} model with a 50\% reduction in \kv. 
}
\vskip -0.1in
\label{fig:attn_dist}
\end{figure}

The natural choice is to utilize the attention score as the score function, denoted as $f_{\theta}(\mathrm{acc;attn})$. This method is commonly observed in previous state-of-the-art work, such as H$_{2}$O~\cite{h2o}. It identifies tokens that consistently receive higher attention scores during the prompt and token generation phases as the most critical or \key.

We can choose and retain these $k$ tokens based on their accumulated attention scores, creating what we refer to as ``Key Attention''. However, relying solely on these $k$ tokens during attention does not provide the necessary accuracy and yields poor performance. This is shown in Figure~\ref{fig:attn_accuracy}.

In this comparison, both `Window Attention' and `Key Attention' demonstrate inferior performance compared to full attention, even when the window and \key parameters are reduced by $\nicefrac{n}{2}$. While reducing the sizes of the window and \key relative to the total tokens ($n$) is crucial for minimizing the size of the \kv, it also leads to a significant decrease in accuracy. This decline primarily stems from the loss of recent context in key-token attention and crucial context in window attention. Building on this observation, we propose a mixed approach that combines selected \key with recent tokens to reduce the \kv size while also preserving accuracy.

\subsubsection{Problem: Uneven  Score Distribution}

Figure~\ref{fig:attn_dist} shows the distribution of attention scores $(f_{\theta})$ for full attention, as described in Equation~\ref{eqn:2}. When \kv is reduced, tokens with lower scores are discarded. This alters the score function, shown in Equation~\ref{eqn:3}, as the term $\sum_{m=n-k}^n e^{x_{m}}$ becomes zero.
\vskip -0.15in
\begin{equation}~\label{eqn:2}
f_{\theta}(x_{i}) = \frac{e^{x_{i}}}{\sum_{j=1}^{n} e^{x_{j}}}, \ \ \ \ i=1,2,\dots,n
\end{equation}
\vskip -0.1in
\begin{equation}\label{eqn:3}
f_{\theta}(x_{i}) = \frac{e^{x_{i}}}{\sum_{j=1}^{k} e^{x_{j}} + \bcancel{\sum_{m=n-k}^{n} e^{x_{m}}}}
\end{equation}
\begin{figure} [ht!]
  \centering
  \includegraphics[width=1\columnwidth]{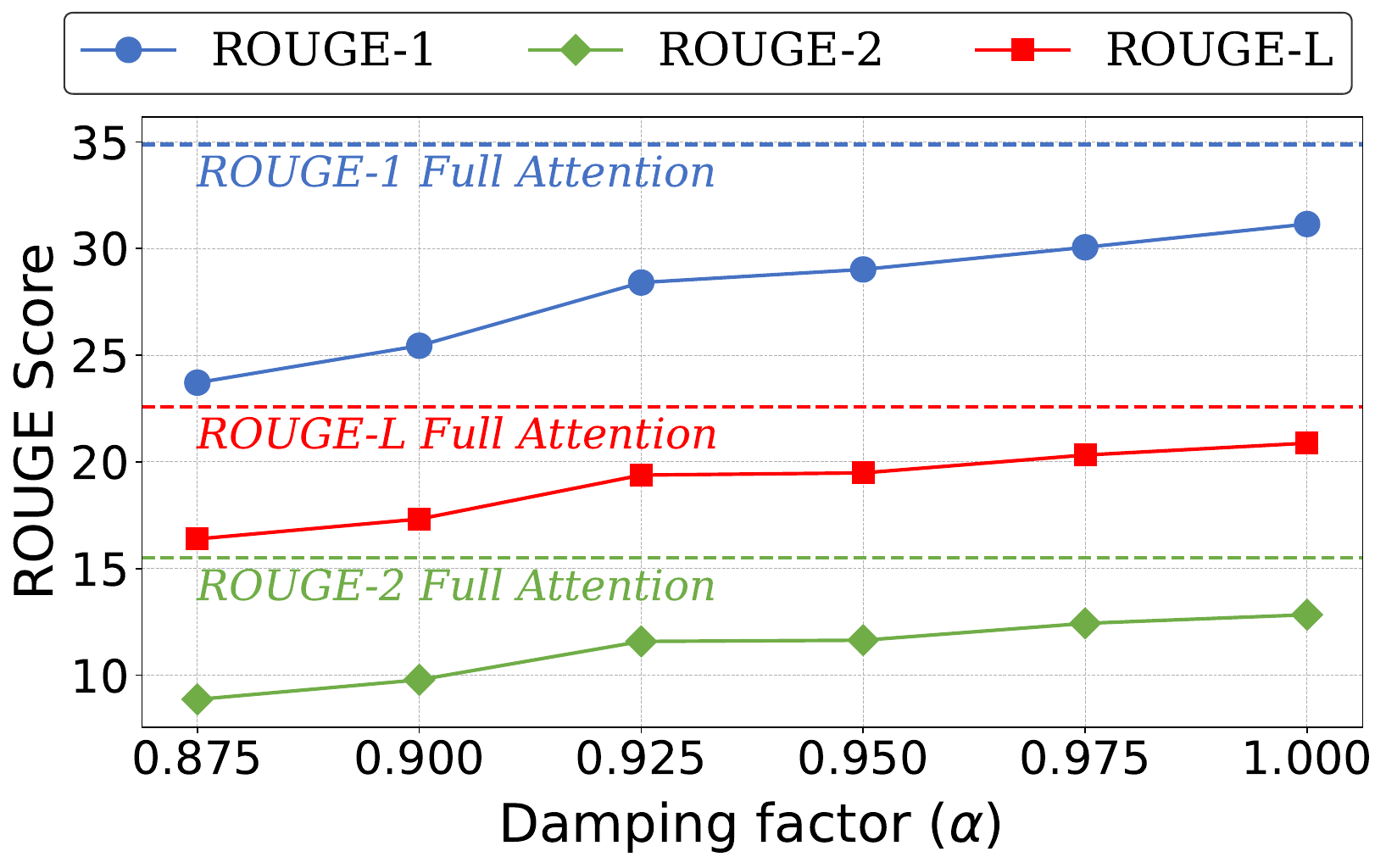}
  \vskip -0.15in
  \caption{The effect of damping on the model quality for Cerebras-GPT-6.7B model with 50\% \kv reduction. Even after damping the score function to counteract the excess attention score, one does not achieve the model quality of the full attention model.}
\vskip -0.2in
\label{fig:damping}
\end{figure}

This removal of tokens disrupts the distribution of the score function. This is because the attention weight of the discarded tokens is unevenly distributed among the tokens within the reduced \kv. This uneven distribution arises due to the nature of the inherent \emph{softmax} function. Figure~\ref{fig:attn_dist} illustrates this phenomenon by comparing the distribution of the score function for full attention with that after \kv reduction. When the score distribution is uneven, the model may not attend to the most relevant tokens in the sequence. Consequently, this can result in a loss of contextual information, reduced accuracy, and potentially lower-quality text generation.

\subsubsection{Motivation: Damping the Score Function}

A straightforward approach involves damping the score function using a damping factor to counteract the excess attention score resulting from discarded tokens. Assume $\alpha$ is the damping factor. It modifies the score function as $\bar{f_{\theta}} = \alpha f_{\theta}$. Ideally, we aim to dampen the score function by a factor equivalent to $\sum_{m=n-k}^n e^{x_{m}}$, where $n-k$ represents the tokens that have been discarded.
 
Figure~\ref{fig:attn_score} shows that, even with a 50\% reduction in the \kv size, the average accumulated attention score of \key remains consistently high, ranging from approximately 90\% to 95\%. We conduct a sweep across a range of values to explore the impact of different damping factors ($\alpha$) on overall model quality. This analysis is performed with a \kv size set at 50\% and a recent ratio of 20\% (representing the percentage of recently generated tokens) for the Cerebras-GPT-6.7B~\cite{cerebrasgpt} model.

However, as depicted in Figure~\ref{fig:damping}, even after the application of a damping factor, it is not possible to achieve the same quality as the full attention model. This discrepancy stems from a significant change in the score distribution of the remaining tokens within the reduced \kv. These findings underscore the inadequacy of relying solely on the accumulated attention score-based score function $f_{\theta}(\mathrm{acc;attn})$ for identifying \key. Hence, addressing the impact of discarded tokens within the score function is crucial to achieve higher model accuracy or meet the accuracy requirements of benchmarks like \emph{MLPerf}.

%% file: body/keyformer.tex
\section{Keyformer: Intuition and Design}
\label{sec:keyformer}

\keyformer leverages the inherent sparsity within decoder layers by identifying \key using a \emph{mixture of recent tokens}. It adjusts the changes in the score function resulting from discarded tokens by applying regularization to the unnormalized logits for the identification of \key.

\subsection{Logits Regularization}
We strategically remove $n-k$ tokens from the context in the prompt processing phase. This helps us maintain a constant \kv size with $k$ tokens during generation and prevents unwarranted memory expansion. Thereafter, \keyformer uses logits regularization technique. The introduction of added distribution ($\zeta$) to regularize the reduced logits enables our model to remain robust and adaptive. It helps identify the \key even in the presence of unknown contexts during inference-time. \keyformer adds this noise to the unnormalized logits derived from $QK^{T}$, as illustrated in Equation~\ref{eqn:4}. Also, the type of distribution added significantly impacts the resulting probability distribution.
\begin{equation}\label{eqn:4}
y_i = x_i + \zeta_i, \ \ \ where\ x_i = \frac{Q[i,:]K[:,i]^{T}}{\sqrt{d_k}}
\end{equation}
Here, $y_i$ are the adjusted logits, $x_i$ are the unnormalized logits, and $\zeta_i$ is the added distribution for regularization.

\subsection{Choice of Distribution for Regularization}
\label{subsec:distribution_choice}

The regularization distribution added to unnormalized logits impacts \key identification and model quality. Thus, we aim to draw intuition using the semantics of LLMs.

\subsubsection{Intuition: Bias Towards Initial Tokens}

Previous research, such as streaming LLMs~\cite{streaming_llm} and the H$_{2}$O model~\cite{h2o}, has shown a bias towards initial tokens. This bias stems from accumulated attention scores favoring initial tokens due to cumulative effects during decoding iterations. We propose using a skewed distribution to leverage this bias and effectively model the distribution of maximum values (\key). This distribution favors initial tokens while maintaining an asymmetric profile, enhancing the representation of tokens drawn from the recent context window.

\vspace{-0.1in}
\paragraph{Gumbel Logit Adjustment:}
Our choice of distribution is inspired by the \emph{Gumbel distribution}~\cite{gumbel}. The Gumbel distribution is particularly well-suited for our \key identification task, as it characterizes the distribution of maximum values within a set of samples and is skewed towards initial tokens. This makes it an ideal candidate for modeling \key for long sequences.

\begin{equation}\label{eqn:7}
f_{Gumbel}(\zeta_{i}) = e^{-\zeta_{i} - e^{-\zeta_{i}}}
\end{equation}
\begin{equation}\label{eqn:8}
f_{Gumbel}(y_i) = e^{-(y_{i}-x_{i}) - e^{-(y_{i}-x_{i})}}
\end{equation}

Equation~\ref{eqn:7} presents the standard Gumbel pdf applied to unnormalized logits, while Equation~\ref{eqn:8} displays the pdf of logits adjusted with Gumbel addition. Additionally, it is noteworthy that the Gumbel distribution holds significance in statistical theory. It captures the essence of the Gumbel limit theorem, which asserts that common probability distributions (such as normal, exponential, uniform, etc.) converge to the Gumbel distribution. This underscores its appropriateness for modeling the identification of \key.

\begin{figure*}[t]
\centering
  \includegraphics[width=1\textwidth]{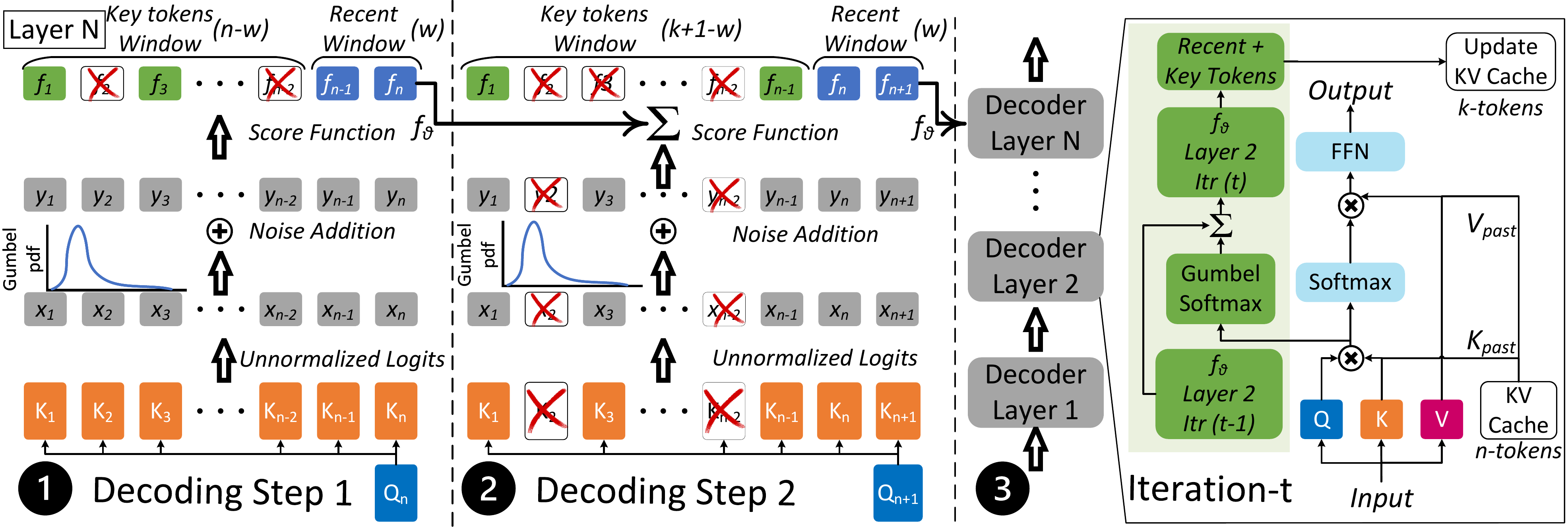}
  \caption{The overview of \keyformer: \myCircled{1} Initial decoding step involves token generation with $n$ tokens in \kv. \keyformer induces noise for \key identification, selecting $w$ tokens from the recent window and $(k-w)$ tokens from the remaining $(n-w)$ to maintain $(k)$ tokens in \kv. \myCircled{2} Subsequent decoding step uses the reduced \kv from the previous iteration. \myCircled{3} The design of \keyformer from the perspective of a single decoder layer involves taking unnormalized logits from $QK^{T}$ and introducing 'Gumbel' noise. This is done using a Gumbel-based probability distribution and helps address the issue of \key being discarded. The score function ($f_{\theta}$) accumulates over decoding steps for each layer and head.}
  \label{fig:keyformer}
  \vspace{-0.2in}
\end{figure*}

In theory, selecting a regularization distribution that promotes uniformity after normalization aids in \key identification. This is crucial during inference when information about discarded tokens is unavailable. To quantify the spread of probability distributions post-normalization, we employ entropy, defined as $\mathit{H}(p) = -\sum p_{i}\log(p_{i})$. Our analysis indicates that Gumbel-based logit adjustment fosters a more uniform distribution, suggesting its effectiveness as a regularization technique for \key identification, as demonstrated in Equation~\ref{eqn:9} and Equation~\ref{eqn:10}.

\begin{equation}\label{eqn:9}
\mathbf{z} = \mathrm{softmax}(\mathbf{y})
\end{equation}

\begin{equation}\label{eqn:10}
\mathit{H}(\mathbb{E} [\mathbf{z}_{Gumbel}])>\mathit{H}(\mathbb{E} [\mathbf{z}])
\end{equation}

\subsection{Keyformer Score Function}

We propose a novel score function for \keyformer, denoted as $f_{\theta}(\mathrm{Keyformer})$, to address the limitations of the accumulated attention-based score function ($f_{\theta}(\mathrm{acc\;attn})$). This new score function integrates the Gumbel noise distribution into the unnormalized logits. However, it fails to account for the discarded tokens in forming the underlying probability distribution. To rectify this, we introduce a temperature parameter, denoted as $\tau$, as shown in Equation~\ref{eqn:11}.
\begin{equation}\label{eqn:11}
f_{\theta}(x_{i}) = \frac{e^{\nicefrac{(x_{i}+\zeta_{i})}{\tau}}}{\sum_{j=1}^k e^{\nicefrac{(x_{j}+\zeta_{j})}{\tau}}}, \ \ \ \ i=1,2,\dots,k
\end{equation}
The probabilistic score functions described above are akin to the concept of \emph{Gumbel Softmax}~\cite{gumbel_softmax}. This score function offers a continuous relaxation of discrete random variables~\cite{concrete_distribution}. This alignment corresponds with our \emph{primary objective of identifying a subset of past tokens} $S_{k} \subset S_{n}$ that conveys the same semantic information as the original complete set of tokens.

\subsubsection{Significance of the Temperature Parameter $(\tau)$}

The `temperature' parameter $(\tau)$ is pivotal in regulating the smoothness of the probabilistic distribution. Higher values of $\tau$ $(\tau \to \infty)$ yield uniform probabilities, assigning equal scores to all tokens. Conversely, lower values of $\tau$ $(\tau \to 0)$ produce a sharper distribution, prioritizing specific tokens based on their unnormalized logits. This parameter governs the degree of randomness in probabilities. It is crucial when tokens are removed from the \kv, as they cannot be reintroduced without recomputing their keys and values.

In Equation~\ref{eqn:12}, we illustrate the dynamic nature of $\tau$ at each decoding iteration $t$. To achieve this, we define a range for $\tau$ spanning from $\tau_{init}$ to $\tau_{end}$. In each decoding step, we increment $\tau$ by $\Delta\tau$, a value determined by the range of $\tau$ and the length of the text being generated, denoted as $T$. 
\begin{equation}\label{eqn:12}
\tau = \tau_{init} + t\Delta\tau, \ \ \ \ \Delta\tau = \frac{\tau_{end} - \tau_{init}}{T}
\end{equation}

This strategy is based on the premise that we need a more uniform or randomized probability distribution as more tokens are discarded. Through empirical analysis, we discovered that setting $\tau_{init}=1$ and $\tau_{end}=2$ produces optimal outcomes (refer to Appendix ~\ref{subsubsec:temperature}). This decision aligns with our objective of maintaining a non-random score function during the prompt phase, where all tokens are available. When $\tau$ is set to one, the Gumbel softmax approach is nearly equivalent to a standard \emph{softmax}. As we advance through decoding iterations and discard more tokens to maintain a static \kv size, we systematically increase the randomness in our score function $f_{\theta}$. This is achieved by incrementally raising $\tau$ with $\Delta\tau$.

\subsubsection{Leveraging Score Function Accumulation}
The accumulation of the score function is essential for identifying \key based on their consistent behavior throughout decoding steps. Without accumulation, token discarding would rely solely on the current token's correlation with previous tokens. Although the correlation of the current token is significant in identifying \key, their behavior should remain consistent across most generated tokens. To discern \key based on this consistent behavior, we accumulate the score function $(f_{\theta})$ across both the prompt and token generation phases, as depicted in Figure~\ref{fig:keyformer}.

\subsection{Keyformer Algorithm}

Figure~\ref{fig:keyformer} presents an overview of \keyformer. We highlight its key functionalities in discarding tokens based on sparsification, using a mixture of recent and \key, and introducing a novel Gumbel softmax-based score function for \key identification. During the prompt processing phase, \keyformer calculates keys and values for all $n$ tokens within the prompt length $S_{n}$ to predict the first token. Given the \kv budget, \keyformer retains a recent window of $w$ recent tokens while discarding $n-k$ tokens from the $n-w$ tokens window, thereby identifying $k-w$ tokens. The top-($k-w$) tokens from the $n-w$ window are selected based on the \keyformer score function. The combination of \key ($k-w$) and recent tokens ($w$) forms the reduced \kv. As there are no discarded tokens during the prompt processing phase, \keyformer uses a temperature parameter, $\tau_{init} = 1$, to approximate the \emph{softmax} probability distribution. This is illustrated in decoding step 1.

\input{body/algorithms/algo}

In the token generation phase, \keyformer operates with a reduced \kv. The first generated token attends solely to the $k$ tokens within the \kv, as depicted in decoding step 2. The recent window $w$ shifts right by a single token, while the score function $f_{\theta}$ accumulates with the score function from the previous decoding step. During each decoding step of the token generation phase, $k-w$ \key are identified from a window of size $k+1-w$. Consequently, one token has been added, and another has been removed from the recent window. Since we add and remove tokens from the `\key' window, we can improve accuracy while maintaining a static \kv size equal to $S_{k}$. Moreover, the temperature parameter $\tau$ increases by $\Delta\tau$ to adjust for the number of removed tokens in the probability distribution of the score function. The detailed algorithm for \keyformer is provided in Algorithm~\ref{alg:keyformer}.
\vspace{-0.1in}

%% file: body/algorithms/algo.tex
\begin{algorithm}[ht!]
   \caption{\keyformer}
   \label{alg:keyformer}
\begin{algorithmic}
   \STATE {\bfseries Input:} \kv size: $k$
   \STATE {\bfseries Input:} Recent Window: $w$
   \STATE {\bfseries Input:} Text Generation Length: $T$
   \STATE {\bfseries Input:} Temperature Parameter: $\tau_{init}$, $\tau_{end}$, $\Delta\tau$
   \STATE {\bfseries Input:} Prompt Sequence Length: $S_{n}$
   \STATE {\bfseries Output:} Reduced Sequence Length: $S_{k}$

   \STATE Initialize $f_{\theta} \gets \phi$ , $\tau \gets \tau_{init}$ , $\Delta\tau \gets \frac{\tau_{end}-\tau_{init}}{T}$
   \STATE Initialize $\zeta_{i} \gets $ \emph{Gumbel Distribution}
   \FOR{$t=0$ {\bfseries to} $T$}
   \STATE $\tau \gets \tau_{init} + t\Delta\tau$
   \IF{phase $\gets$ prompt}
   \STATE $x_{i} 
\gets\frac{Q_{i}K_{S_{n}}^{T}}{\sqrt{d}}, m \gets n$
   \ELSE
   \STATE $x_{i} 
\gets\frac{Q_{i}K_{S_{k}}^{T}}{\sqrt{d}}, m \gets k$
   \ENDIF
   \STATE $f_{\theta}(i) \gets f_{\theta}(i) \; + \; \frac{e^{\nicefrac{(x_{i}+\zeta_{i})}{\tau}}}{\sum_{j=1}^m e^{\nicefrac{(x_{j}+\zeta_{j})}{\tau}}}$
   \STATE $S_{w} \gets $ \emph{Recent $w$ tokens}
   \STATE $S_{key} \gets \mathrm{arg}\max_{(k-w)} f_{\theta}[\;:-w]$
   \STATE $S_{k} \gets S_{key} \cup S_{w}$
   \ENDFOR
\end{algorithmic}
\end{algorithm}
\vspace{-0.1in}

%% file: body/evaluation.tex
\begin{figure*}
  \centering
  \includegraphics[width=1.85\columnwidth]{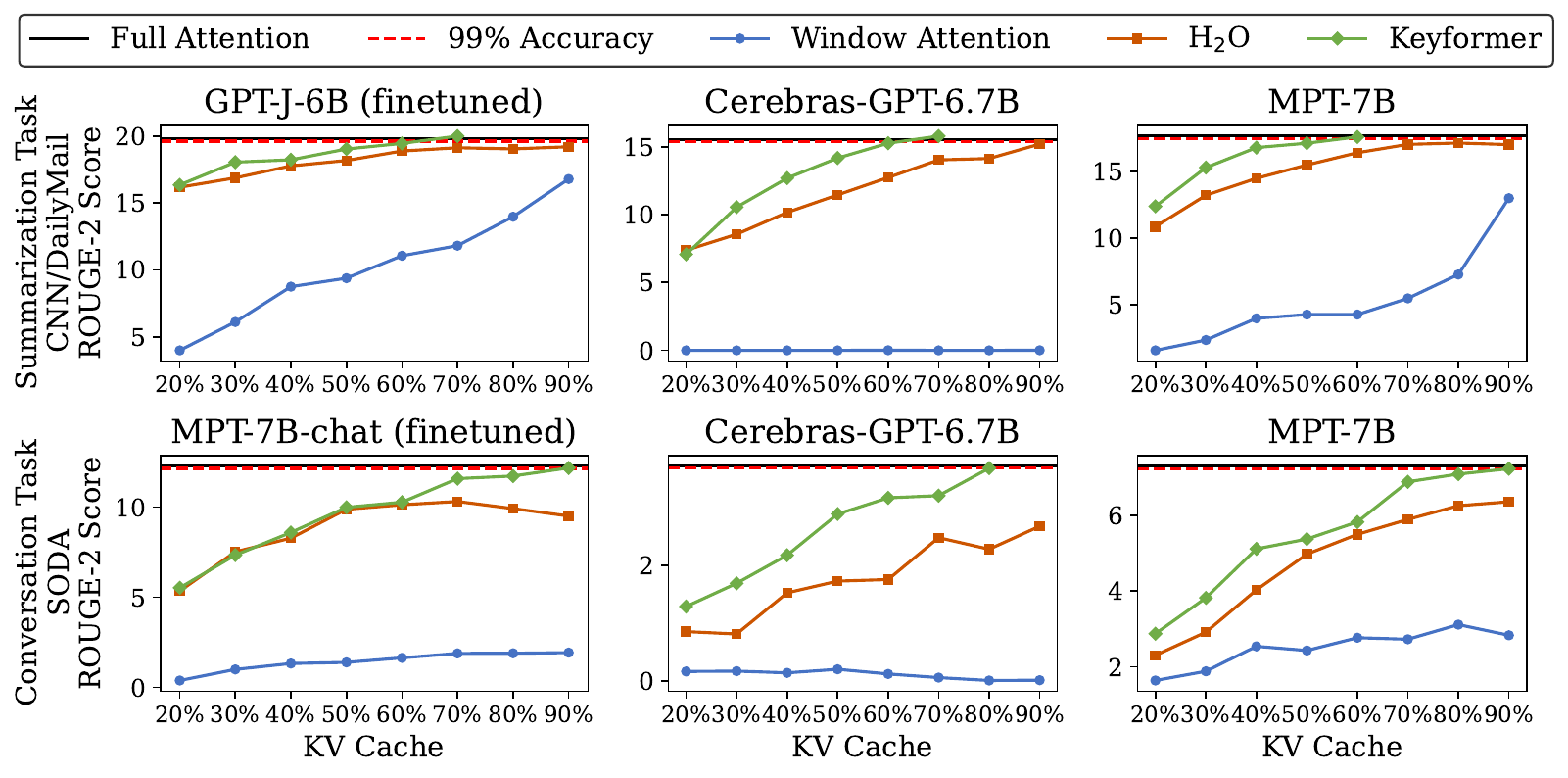}
  \vskip -0.15in
  \caption{The accuracy comparison involves Full Attention, Window Attention, H$_{2}$O, and \keyformer with different \kv sizes. The solid black line represents Full Attention without discarding tokens and with a full \kv. The red dotted line denotes the 99\% accuracy threshold, aligning with the MLPerf guidelines~\cite{mlperf}. Despite using 90\% \kv, both Window Attention and H$_{2}$O fall short of the desired accuracy. In contrast, \keyformer achieves baseline accuracy with only 70\% of the \kv size.
}
\label{fig:accuracy}
\vskip -0.15in
\end{figure*}

\section{Evaluation}
\label{sec:evaluation}

We evaluate \keyformer across three significant model families: GPT-J~\cite{gptj}, Cerebras-GPT~\cite{cerebrasgpt}, and MPT~\cite{mpt}, each using distinct position encoding techniques. GPT-J incorporates RoPE~\cite{rope}, Cerebras-GPT employs learnable position embeddings, and MPT utilizes ALiBi~\cite{alibi}. By including models with varied position encoding methods, we ensure the robustness and generalizability of our findings across representative model families. We employed a fixed beam size of 4 for all evaluations.

\paragraph{Setup:} We conducted evaluations on two representative text generation tasks: summarization, utilizing the CNN/DailyMail~\cite{cnn_dm} and GovReport~\cite{gov_reports} datasets, and conversation, employing the SODA dataset~\cite{soda}. The GPT-J model was fine-tuned specifically for summarization, while Cerebras-GPT and MPT are pre-trained models. We utilized the MPT-chat version of the MPT model for conversation tasks, which was fine-tuned for dialogue generation. All models were pre-trained with a sequence length of 2k.

To address long document summarization, we utilized the MPT-storywriter version of the MPT model, fine-tuned for writing fictional stories. This model accommodates a context length of 65k and can generate content up to 84k tokens long. Additionally, we evaluated four tasks from the lm-eval-harness~\cite{lm_eval} framework: PIQA~\cite{piqa}, Winogrande~\cite{winogrande}, OpenBookQA~\cite{openbookqa}, and COPA~\cite{copa}. These tasks involve few-shot evaluation of autoregressive language models and were executed using the NVIDIA A100 (80GB) GPUs.

\paragraph{Baselines:} To evaluate the accuracy of \keyformer, we compared it against Full Attention. Full Attention acts as our benchmark and represents the gold standard for accuracy. We aim to achieve an accuracy target within the range of 99\% to 99.9\% of Full Attention. This goal aligns with the high-quality standards set by industry benchmarking entities like MLPerf~\cite{mlperf}. Additionally, we performed comparisons with Window Attention and the recent H$_{2}$O model~\cite{h2o}, adjusting the \kv size from 20\% to 90\% of the prompt length.

\subsection{Accuracy Results}
\label{subsec:accuracy}

To assess the impact of \kv reduction on text generation quality, we relied on the ROUGE score~\cite{rouge}, a widely-used metric for evaluating fluency and coherence. ROUGE measures the overlap of n-grams between generated and reference text, providing a standardized measure for text quality. According to MLPerf, ROUGE scores, including \emph{ROUGE-1, ROUGE-2, and ROUGE-L}, should reach 99\% to 99.9\% of their original values for summarization tasks. Thus, even with reduced \kv, our model should maintain the desired ROUGE scores. Figure~\ref{fig:accuracy} depicts accuracy comparisons between \keyformer and other methods (Full Attention, Window Attention, and H$_{2}$O) across different \kv sizes. The illustration focuses on the ROUGE-2 score, which measures bi-gram overlap. Trends for ROUGE-1 and ROUGE-L are detailed in Appendix~\ref{subsec:rouge_r}.

The results highlight the importance of the previous context for model performance. For instance, Window Attention relies solely on recent tokens, leading to a significant loss of accuracy. Thus, identifying \key is crucial for achieving desired model accuracy. Across various \kv budgets, \keyformer consistently outperforms the state-of-the-art H$_{2}$O. It shows that the \key it identifies are more important than the heavy hitters identified by H$_{2}$O. For instance, \keyformer attains the target ROUGE score with just 70\% of the \kv, whereas H$_{2}$O fails to reach this goal even with a larger \kv budget. Furthermore, \keyformer surpasses the baseline accuracy by up to 1.73\% (0.9\% for GPT-J-6B and 1.73\% for Cerebras-GPT-6.7B for summarization task) achieved with full attention. This demonstrates the regularization effect of the introduced Gumbel noise in the score function of \keyformer and its positive impact on \key identification.

\begin{figure}[htbp]
  \centering
  \includegraphics[width=0.75\columnwidth]{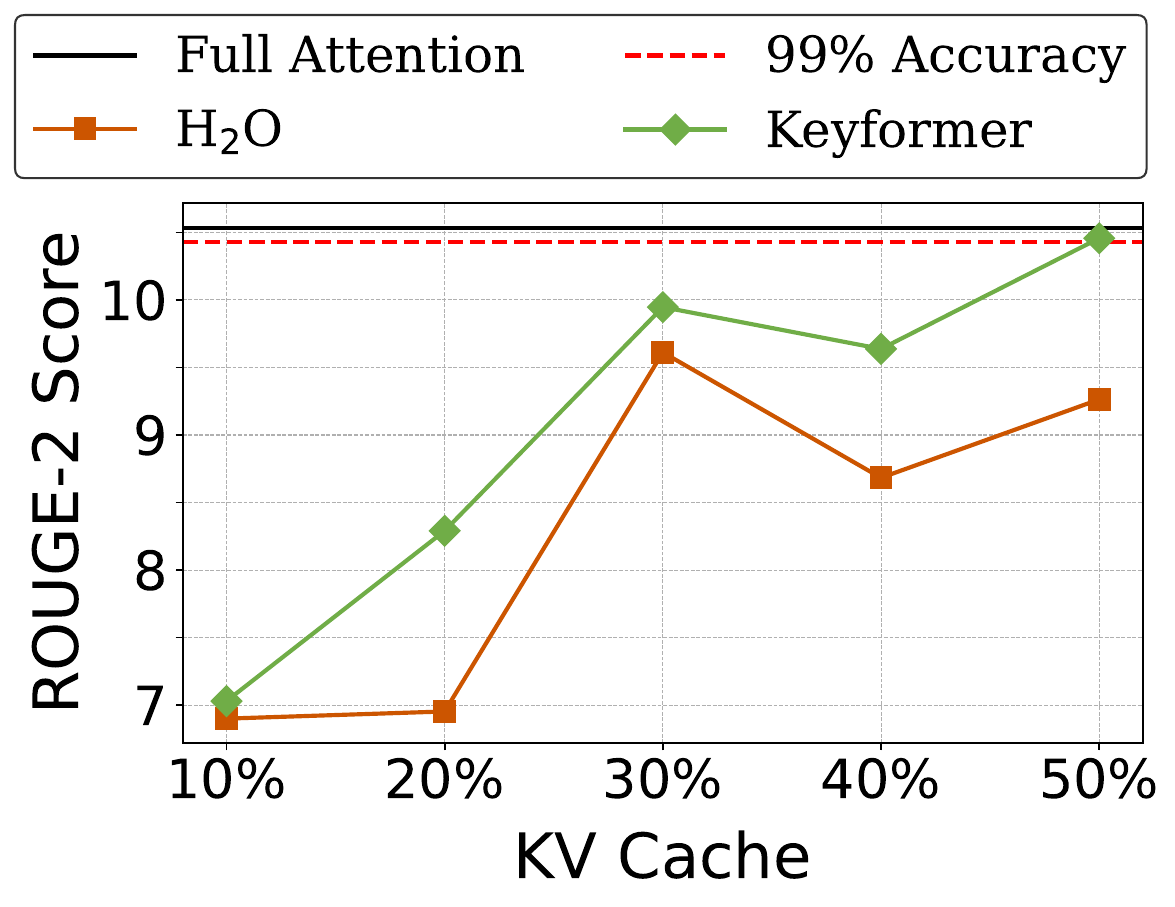}
  \vskip -0.15in
  \caption{Evaluating long context summarization using MPT-7B-storywriter for GovReport dataset with 8k sequence length. 
}
\label{fig:long_context}
\vskip -0.2in
\end{figure}

\paragraph{Long Context Summarization:} We assessed the effectiveness of \kv reduction in \keyformer while maintaining accuracy for handling long contexts. This evaluation was conducted on the MPT-7B-story writer model, pre-trained with a context length of 65k. We utilized the Government report~\cite{gov_reports} dataset, which contains reports authored by government research agencies and features longer summaries and documents. This dataset requires a deep understanding of context to extract crucial information for summarization. Figure~\ref{fig:long_context} shows the accuracy comparison among \keyformer, H$_{2}$O, and Full Attention. Notably, even with a 50\% \kv size, \keyformer maintains the desired 99\% accuracy threshold, while H$_{2}$O shows significantly lower accuracy at the same \kv size.

\subsection{Performance Results}
\label{subsec:performance}

\begin{figure}[t]
  \centering
  \includegraphics[width=0.9\columnwidth]{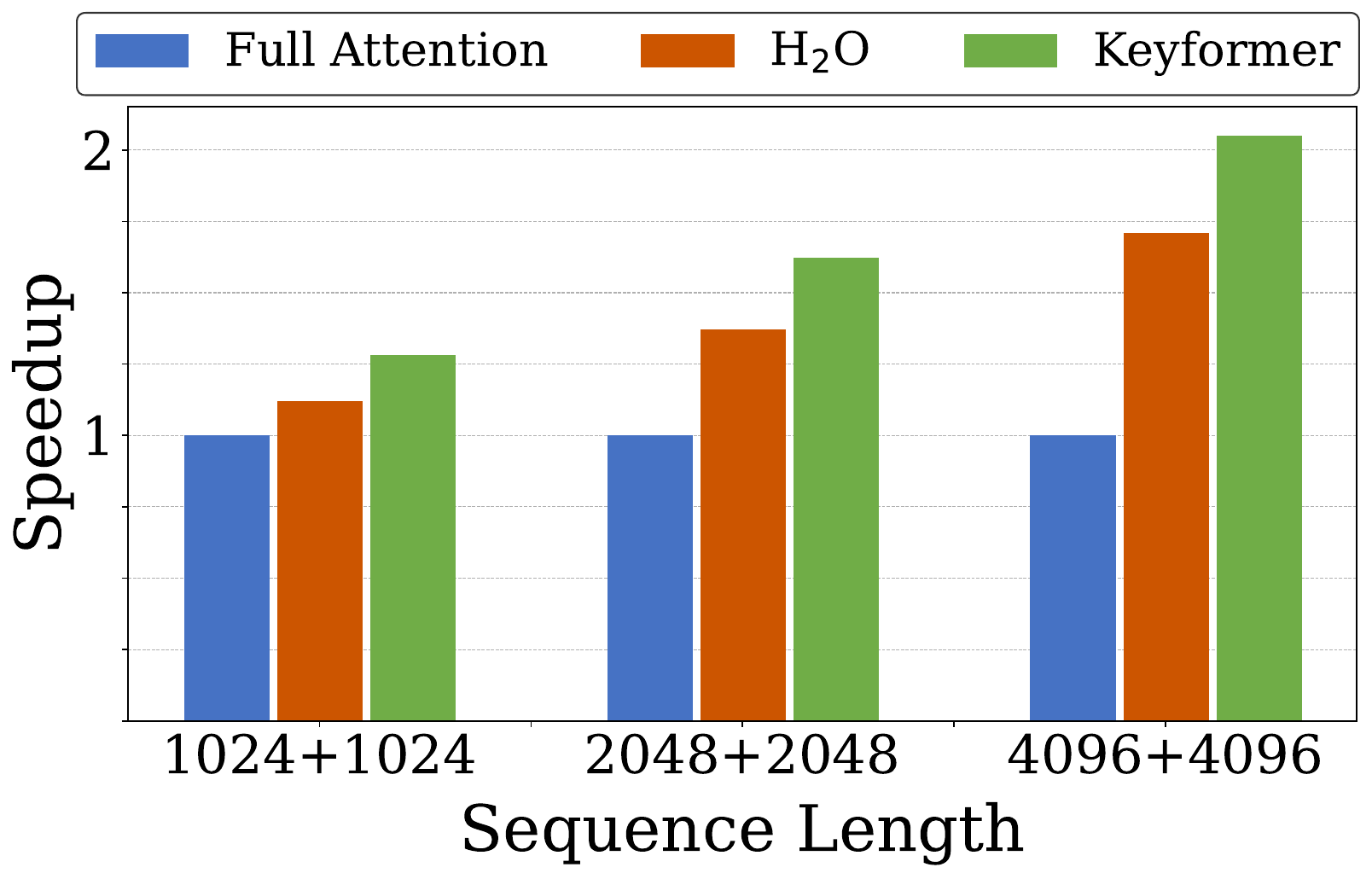}
  \vskip -0.15in
  \caption{Speedup of inference in an iso-accuracy setting for the MPT-7B model. Here, \keyformer reduces \kv by 50\% and H$_{2}$O reduces \kv by 10\%. H$_{2}$O falls short of baseline accuracy with a 50\% \kv.
}
\label{fig:speedup}
\vskip -0.1in
\end{figure}

To assess the performance advantages of \keyformer with reduced \kv, we considered two critical inference metrics: inference latency and generation throughput for the target models. Our \keyformer implementation is seamlessly integrated with Huggingface~\cite{huggingface} model cards, ensuring ease of adoption. We disabled CPU offloading in cases where the model and \kv exceeded GPU HBM memory capacity, ensuring consistent evaluation. We generated a synthetic dataset to maintain evaluation consistency, where all prompts were padded with synthetic text. We employed the MPT-7B-storywriter model to generate an equal number of tokens for each prompt. We tested various combinations of prompt and generation lengths.

Figure~\ref{fig:speedup} presents inference latency speedup while Table~\ref{tab:performance} shows improvement in generation throughput in comparison to a Full Attention-based method. With a 50\% \kv reduction, \keyformer significantly reduces inference latency, achieving \latencyimprv with the same batch size. Moreover, the reduced \kv size allows \keyformer to handle twice the batch size compared to full attention, increasing token generation throughput by \throughputimprvsame with the same batch size and \throughputimprv with a bigger batch size.

\input{body/tables/throughput}

\paragraph{Performance Improvement Breakdown:}

To understand the sources of performance benefits with \keyformer-based \kv reduction, we primarily consider two factors:
\setlist{nolistsep}
\begin{enumerate}[leftmargin=*,noitemsep]
    \item \textbf{Reduced $\mathsf{\textbf{KV}}$ $\mathsf{\textbf{cache}}$}: A smaller \kv significantly reduces the data movement from off-chip GPU HBM.
    \item \textbf{Scaled Dot Product Optimization}: The number of tokens in the \kv is reduced from $n$ to $k$.
\end{enumerate}

The above two factors reduce the overall smaller size of matrices. Thus, they enable an optimized scaled dot product within the multi-head attention block $(QK^{T})V$.

\begin{figure}[t]
  \centering
  \includegraphics[width=1\columnwidth]{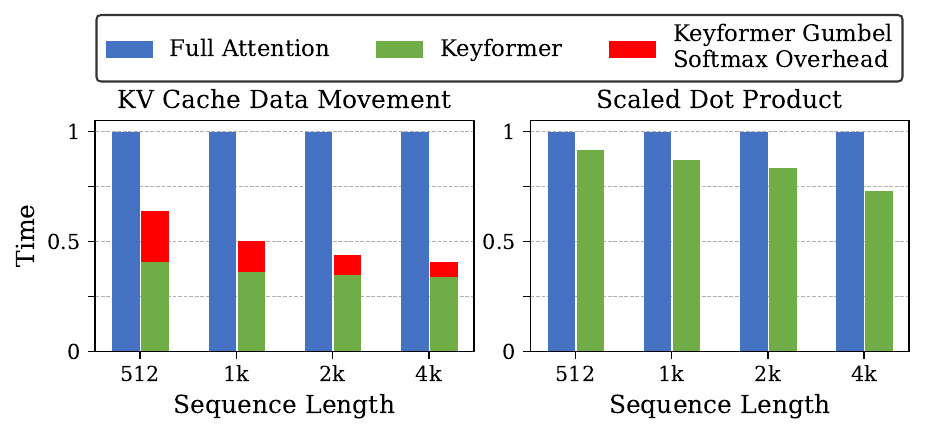}
  \vskip -0.15in
  \caption{Normalized time for \kv data movement and Scaled Dot Product $(QK^{T})V$ with \keyformer 50\% \kv reduction along with its score function overhead. 
}
\label{fig:performance}
\vskip -0.2in
\end{figure}

It is worth noting that in LLMs, which are memory-bound, the main performance boost comes from reducing \kv data movement rather than matrix multiplication. However, \keyformer's \key identification process introduces some overhead due to Gumbel softmax. Figure~\ref{fig:performance} illustrates the normalized performance improvement for \keyformer, considering both reduced \kv data movement and optimized scaled dot product. These enhancements are demonstrated for the MPT-7B-storywriter model with a 50\% \kv reduction, including the additional overhead from \keyformer's Gumbel softmax. The results indicate that \keyformer-based \kv reduction decreases \kv data movement by \kvdatareduction and improves computational efficiency in the attention module's scaled dot product by \matmulimprv, particularly for sequences of length 4k.

\subsection{Few-Shot Evaluation}
\label{subsec:few_shot}

We performed few-shot experiments using four tasks from the \texttt{lm-eval-harness} framework and pre-trained models to evaluate \keyformer's performance under varying numbers of shots during inference. Table~\ref{tab:few_shot} presents the results for 0 and 5 shots, demonstrating that \keyformer consistently surpasses previous approaches across all tasks and shot settings. Even with a 50\% reduction in \kv size, it achieves accuracy close to the full attention baseline.

\input{./body/tables/lm_eval}

\input{body/tables/ablation_accuracy}

\subsection{Ablation Studies}
\label{subsec:ablation}

\subsubsection{Shared versus Per-Layer Score Function}

The score function $f_{\theta}$ defines what constitutes a \key in the context. In generative LLMs with stacked decoder layers, the score function ($f_{\theta}$) can either be shared across all layers (\emph{Shared}) or dedicated to each layer (\emph{Per-Layer}). In the \emph{Per-Layer} approach, $f_{\theta}$(Per-Layer) assigns a dedicated score function to each decoder layer, with accumulation occurring at each decoding stage. Conversely, $f_{\theta}$(Shared) uses a single global score function for all decoder layers, with accumulation across decoder layers and decoding stages. 

Table~\ref{tab:ablation_accuracy} illustrates the accuracy comparison between Per-Layer and Shared score functions, maintaining the original positional information and \kv size constant. Notably, using the Per-Layer score function yields better accuracy than the shared score function. This aligns with the intuition that transformers learn hierarchical text representations across layers, with lower layers capturing local syntactic and semantic features and higher layers capturing more abstract and complex patterns~\cite{hiclip}. Therefore, having a Per-Layer score function for each layer aids in \key identification specific to that layer.

\subsubsection{New vs. Original Positional Information}

We investigated how reducing the \kv size affects the positional information used for the keys in \keyformer. We explored two approaches: \emph{\keyformer(Org Pos)} and \emph{\keyformer(New Pos)}. In \emph{\keyformer(Org Pos)}, the position information reflects the original positions of tokens within the text. Conversely, \emph{\keyformer(New Pos)} uses positions based on the new arrangement of tokens within \kv.

Table~\ref{tab:ablation_accuracy} presents the accuracy comparison with consistent \kv size and score function. Notably, when using the original positional information, \keyformer excels in accuracy. However, incorporating new positional information during inference leads to a slight drop in accuracy. Nevertheless, even with new positional information, \keyformer outperforms the state-of-the-art H$_{2}$O.

\subsubsection{Comparison with Alternative Distributions}
\label{subsubsec:logit_adjustment}

We conducted an ablation study to assess how different logit adjustment distributions affect model accuracy or \key identification. We evaluated three regularization strategies and compared them with Gumbel-based logit adjustment. These are no logit adjustment, constant logit adjustment, and Gaussian distribution-based logit adjustment.

\vspace{-0.05in}
\paragraph{No Logit Adjustment:}
To examine the effect of omitting regularization on unnormalized logits, we experimented without logit adjustment, where $y_i = x_i + \bcancel{\zeta_i}$, mirroring the approach used in H$_{2}$O~\cite{h2o}.

\vspace{-0.05in}
\paragraph{Constant Logit Adjustment:}
To study the impact of constant regularization on all unnormalized logits, we experimented with constant logit adjustment, setting $y_i = x_i + c$, where $c$ is the constant that is being added to every unnormalized logit.

\vspace{-0.05in}
\paragraph{Gaussian Logit Adjustment:}
We also used a symmetric Gaussian distribution for logit adjustment.
\begin{equation}\label{eqn:5}
f_{Gaussian}(\zeta_{i}) = \frac{1}{\sqrt{2\pi\sigma^{2}}}\exp\left(-\frac{(\zeta_i - \mu))^2}{2\sigma^2}\right)
\end{equation}
\begin{equation}\label{eqn:6}
    f_{Gaussian}(y_{i}) = \frac{1}{\sqrt{2\pi\sigma^{2}}}\exp\left(-\frac{(y_i - (x_i + \mu))^2}{2\sigma^2}\right)
\end{equation}
Equation~\ref{eqn:5} presents the Gaussian probability density function (pdf) with mean $\mu$ and variance $\sigma^2$ applied to unnormalized logits, while Equation~\ref{eqn:6} displays the pdf of logits adjusted with Gaussian addition.

We established the baseline accuracy lower bound for Gumbel-based adjustments after analyzing the accuracy of the summarization task on the CNN/Daily Mail dataset. We evaluated this task with a 60\% reduction in \kv. Table~\ref{tab:logit_adjustment} provides a comparison of different approaches. We utilized a standard Gumbel pdf with $\mu=0.5772$ and $\sigma=1.2825$. For comparison, the Gaussian pdf had an identical mean and variance, and the constant logit adjustment employed a constant value of $c=0.5772$. The ``No logit adjustment'' approach uses the method in prior work, H$_{2}$O.

Thus, empirical evidence shows that the Gumbel distribution, known for its skewness to initial tokens, is an effective regularization mechanism for $\mathsf{key}$ $\mathsf{token}$ identification.

\input{./body/tables/logit_adjustment}

\subsubsection{Recent Window versus Key token Window Ratio}

We conducted a sensitivity study to examine the impact of varying the ratio of recent tokens $w$ on the size of the \kv. This resulted in changes in the number of \key $(k-w)$. Results in Appendix~\ref{subsec:recent_vs_key_ratio} indicate that the models perform better when the recent tokens ratio $w$ falls within the range of 20\% to 30\%. This observation aligns with our hypothesis that recent and \key are critically important for LLM inference.

\subsubsection{Comparison with Attention Sinks}
Recent research introduced StreamingLLM~\cite{streaming_llm}, which introduced the concept of ``attention sinks.'' StreamingLLM enables Language Models (LLMs) trained with a finite-length attention window to handle infinite sequence lengths without fine-tuning. This is achieved by retaining the first four tokens (known as ``attention sinks'') and a moving window of recent tokens, where $w = k-4$. To compare StreamingLLM with \keyformer, we maintained a \kv size of 60\% for both techniques. Table~\ref{tab:ablation_accuracy} displays the accuracy comparison, showing that StreamingLLM struggles in summarizing text by relying on only the first four tokens as attention sinks and the remaining tokens from a recent window (Appendix~\ref{subsubsec:attn_sink}).

%% file: body/tables/throughput.tex
\begin{table}[t]
  \caption{The generation throughput (tokens/sec) for the MPT-7B model across different sequence lengths. ``1024 + 1024'' shows the sum of the prompt length and the token generation length. ``OOM'' stands for out-of-memory and ``BS'' for batch size.}
  \label{tab:performance}
  \centering
  \resizebox{1\columnwidth}{!}{
  
  \begin{tabular}{l c c c}
    \toprule
    \multirow{2}{*}{\textbf{Sequence Length}} & \textbf{Full Attention} & \textbf{H$_{2}$O} & \textbf{\keyformer}\\
     & Original cache & 90\% \kv & 50\% \kv \\
     \midrule
     1024 + 1024 & 24.9 & 27.8 & 32.0 \\
    2048 + 2048 & 15.0 & 20.5 & 24.3 \\
    4096 + 4096 (BS=1) & 8.3 & 14.1 & 17.0 \\
    4096 + 4096 (BS=2) & OOM & OOM & 19.85 \\
    \bottomrule
  \end{tabular}}
  \vspace{-0.15in}
\end{table}

%% file: body/tables/lm_eval.tex
\begin{table}
  \caption{Few shot results for different tasks. H$_{2}$O and \keyformer are with 50\% \kv.}
  \label{tab:few_shot}
  \centering
  \resizebox{0.95\columnwidth}{!}{
  \begin{tabular}{l |c |c c| c c}
    \toprule
    \multirow{2}{*}{\textbf{Task}} & \textbf{Attention} &  \multicolumn{2}{c|}{\textbf{Cerebras-GPT-6.7B}} & \multicolumn{2}{c}{\textbf{MPT-7B}} \\
     & \textbf{Method} & \textbf{0-Shots} & \textbf{5-Shots} & \textbf{0-Shots} & \textbf{5-Shots} \\
    \midrule
    \multirow{3}{*}{COPA} &  Full & 73.0 & 73.0 & 80.0 & 83.0  \\
    & H$_{2}$O & 68.0 & 74.0 & 75.0 & 82.0  \\
    & \keyformer & 70.0 & 74.0 & 76.0 & 84.0  \\
    \midrule
    \multirow{3}{*}{OpenBookQA} &  Full & 34.8 & 36.8 & 41.8 & 43.4  \\
    & H$_{2}$O & 32.2 & 38.0 & 37.6 & 44.0  \\
    & \keyformer & 32.6 & 38.6 & 40.6 & 43.2  \\
    \midrule
    \multirow{3}{*}{Winogrande} &  Full & 60.2 & 58.4 & 68.7 & 71.8  \\
    & H$_{2}$O & 56.7 & 59.5 & 63.2 & 72.1  \\
    & \keyformer & 57.9 & 59.7 & 64.1 & 72.3  \\
    \midrule
    \multirow{3}{*}{PIQA} &  Full &74.2- & 74.4 & 79.9 & 80.7  \\
    & H$_{2}$O & 72.9 & 73.9 & 79.4 & 79.8  \\
    & \keyformer & 73.0 & 73.6 & 79.2 & 80.1  \\
    \bottomrule
  \end{tabular}}
  \vskip -0.2in
\end{table}

%% file: body/tables/ablation_accuracy.tex
\begin{table*}[t]
  \caption{ROUGE Score comparison with different methods and score functions for summarization task using the CNN/DailyMail dataset.}
  \label{tab:ablation_accuracy}
  \centering
  \resizebox{1.35\columnwidth}{!}{
  \begin{tabular}{l c c c c c c}
    \toprule
    \multirow{2}{*}{\textbf{Model}} & \textbf{Attention} & \textbf{Score fn} & \textbf{KV Cache} & \multirow{2}{*}{\textbf{ROUGE-1}} & \multirow{2}{*}{\textbf{ROUGE-2}} & \multirow{2}{*}{\textbf{ROUGE-L}}\\
     & \textbf{Method} & \textbf{$f_{\theta}$} & \textbf{Size} & & & \\
    \midrule
    \multirow{8}{*}{MPT-7B} & Full & - & Original & 38.6373 & 17.6329 & 24.506 \\
     & Full (99\% Accuracy) & - & Original & 38.2509 & 17.4565 & 24.2609 \\
    & Window & - & 60\% & 18.1296 & 4.2655 & 11.5288 \\ 
     & H$_{2}$O & Per-Layer & 60\% & 36.9616 & 16.3865 & 24.2301 \\
     & StreamingLLM & - & 60\% & 1.3572 & 0.0179 & 1.0281 \\
     & \emph{\keyformer(New Pos)} & Per-Layer & 60\% & 36.9152 & 16.9092 & 23.7218\\
     & \textbf{\emph{\keyformer(Org Pos)}} & Per-Layer & \textbf{60\%} & \textbf{38.7134} & \textbf{17.5976} & \textbf{24.5724} \\
     & \emph{\keyformer(Org Pos)} & Shared & 60\% & 38.2537  & 17.3732  & 24.2579  \\
    \bottomrule
  \end{tabular}}
   \vskip -0.2in
\end{table*}

%% file: body/tables/logit_adjustment.tex
\begin{table}[h!]
  \vskip -0.1in
  \caption{Empirical evaluation with different logit adjustments. Summarization task with 60\% \kv.}
  \label{tab:logit_adjustment}
  \centering
  \resizebox{1\columnwidth}{!}{
  \begin{tabular}{l |c c c c}
    \toprule
    \multirow{2}{*}{\textbf{Model}} & \multicolumn{4}{c}{\textbf{ROUGE-2 Score}} \\
     & \textbf{Gumbel} & \textbf{Gaussian} & \textbf{Constant} & \textbf{None} \\
    \midrule
    GPT-J-6B &  \textbf{19.44} & 14.53 & 12.49 & 18.87 \\
    Cerebras-GPT-6.7B &  \textbf{15.25} & 9.54 & 8.98 & 12.73  \\
    MPT-7B &  \textbf{17.57} & 10.17 & 7.56 & 16.38  \\
    \bottomrule
  \end{tabular}}
  \vskip -0.1in
\end{table}

%% file: body/related_work.tex
\section{Related Work}
\label{sec:related_work}

\paragraph{Attention Speedup:} Prior work focus on improving inference speed for transformer~\cite{attention} based models. PoWER-BERT~\cite{goyal2020power} utilizes word-vector elimination by exploiting redundancy for encoder-based models. Linformer~\cite{linformer} tries to reduce the attention mechanism from quadratic to linear. Reformer~\cite{reformer} reduces attention complexity by locality-sensitive hash (LSH). Linear transformers~\cite{katharopoulos2020transformers} store accumulated states rather than preserving every representation. FLAT~\cite{kao2023flat} suggests optimized dataflow, while other research~\cite{wang2022overlap} overlaps communication with dependent computation to enhance attention execution. In contrast, \keyformer aims to reduce \kv, speeds up attention by reducing the tokens.

\paragraph{Sparse Attention:} One line of work sparsifies the attention mechanism to reduce the computational and memory capacity of the attention block. BigBird~\cite{bigbird} combines random, windowed, and global attention to maintain the accuracy for transformers while sparsifying the attention block. LongFormer~\cite{longformer} also utilizes windowed attention with task-based local attention to achieve sparse attention. Spatten~\cite{spatten} introduces sparsity at both the head and token levels. However, it needs a dedicated architecture to exploit sparsity. Furthermore, these works do not address inference optimizations.

\paragraph{KV Cache Reduction:} El-Attention~\cite{el-attention} modifies the multi-head attention module to reduce the \kv size, leveraging key and value stability during incremental decoding for reuse across layers. In contrast, H$_{2}$O~\cite{h2o} identifies heavy-hitters and keeps them in the \kv to reduce its size, neglecting the attention score distribution shift that occurs post elimination of previous tokens from the \kv, leading to accuracy trade-offs. Other approaches~\cite{deja, dynamic_ctx_pruning} introduce sparsity at both coarse and fine-grained levels, targeting the elimination of specific heads and tokens during inference. However, these methods require task-specific predictors and fine-tuning of pre-trained models. Another method~\cite{gisting} compresses prompts into gist tokens to reduce \kv. Landmark Attention~\cite{landmark} represents token blocks with an additional landmark token in the vocabulary, necessitating computationally intensive retraining or fine-tuning for gist or landmark token integration.

%% file: body/future_work.tex
\section{Future Work}
\label{sec:future_work}

Recent techniques like Multi-Query Attention (MQA)~\cite{mqa} and Group-Query Attention (GQA)~\cite{gqa} aim to train foundation models with fewer attention heads. 
However, such models are typically used after fine-tuning for specific tasks. 
While the detailed evaluation of \keyformer with these models is deferred to future work, it is worth noting that \keyformer can still be applied on top of MQA or GQA-based models. This is because it discards redundant tokens regardless of the number of heads. Additionally, we plan to integrate \keyformer into the LLM's attention block by replacing the standard softmax with a \keyformer-based softmax. This introduces sparsity during training, addressing the quadratic computational and memory complexities of transformers.
This direction aims to enhance scalability to longer contexts without sacrificing accuracy. 

%% file: body/conclusion.tex
\section{Conclusion}
\label{sec:conclusion}

Advancements in large language models (LLMs) are pushing for longer contexts and extensive text generation, with models trained on sequences of millions of tokens. However, this trend strains system memory bandwidth, leading to execution costs. In longer contexts, the \kv size, primarily responsible for memory bandwidth consumption and inference latency, exceeds the model parameters' size. To address this, we proposed \keyformer, which effectively reduces the \kv size upto 50\% without sacrificing accuracy by discarding tokens across heads, layers, and beams, identifying essential tokens (\key) based on a novel score function. \keyformer can be applied to LLMs at inference time, without requiring fine-tuning, while also improving latency and token generation throughput.

\section*{Acknowledgements}

We thank the anonymous reviewers for their feedback. Muhammad Adnan's Ph.D. is supported by the Intel TSA and Natural Sciences and Engineering Research Council of Canada (NSERC) [RGPIN-2019-05059] Grants.

%% file: body/appendix.tex
\section{Appendix}
\subsection{Qualitative Comparison of Text Generation}
\label{subsec:qualitative}

This section qualitatively and quantitatively compares text generation during summarization tasks using various \kv reduction methods on the MPT-7B~\cite{mpt} pre-trained model, focusing on \keyformer. The qualitative assessment is based on a random sample from the CNN/DailyMail~\cite{cnn_dm} validation dataset, and ROUGE scores~\cite{rouge} are employed for quantitative comparisons. \keyformer performs better than other techniques with similar \kv reduction. 

\finding{\textbf{\underline{Input:}} Douglas Costa will spark a transfer scramble this summer with Shakhtar Donetsk ready to sell their prized-asset. Chelsea manager Jose Mourinho is a known admirer of the Brazil international having tried to land the midfielder in the previous transfer windows. Shakhtar chiefs are now open to selling Costa this summer and talks with third parties over his departure are underway. Brazil international Douglas Costa is set to depart Shakhtar Donetsk for £25million in the summer . Midfielder Costa (left) could spark a bidding war from Chelsea, Real Madrid and Barcelona . The 24-year-old Costa wants to remain playing for a Champions League side and is open to a move to England . And the Ukrainian side hope to spark a bidding war for the midfielder in attempt to maximise his transfer fee, with a starting price of £25million. Costa is open to joining a Barclays Premier League club, but wants to be playing Champions League football. However, a host of Europe's top sides will also be in the running to land Costa this summer, Real Madrid and Barcelona among them. Arsenal have had Costa watched extensively, while Manchester United and Liverpool have both been linked with a move. Chelsea manager Jose Mourinho is known to be an admirer of the midfield player Costa . Costa has dual Brazilian and Portuguese nationality, removing any potential work permit problems . Chelsea are set to lead the charge for the 24-year-old having tracked the Brazil international for two years. Despite playing for Brazil, Costa has dual Portuguese nationality so should not face work permit issues.
}

\finding{\textbf{\underline{Reference Summary:}} Shakhtar Donetsk could sell Douglas Costa for at least £25million. Jose Mourinho is a known admirer of the Brazil international midfielder. Real Madrid, Barcelona and Arsenal are also interested in the 24-year-old. READ: Douglas Costa dreaming of summer move to Chelsea .
}

\finding{\textbf{\underline{MPT-7B Full Attention:}} Douglas Costa will spark a transfer scramble this summer with Shakhtar Donetsk ready to sell their prized-asset. Chelsea manager Jose Mourinho is a known admirer of the Brazil international having tried to land the midfielder in the previous transfer windows. Shakhtar chiefs are now open to selling Costa this summer and talks with third parties over his departure are underway. Brazil international Douglas Costa is set to depart Shakhtar Donetsk for £25million in the summer . Midfielder Costa (left) could spark a bidding war from Chelsea, Real Madrid and Barcelona .\\
\textbf{ROUGE-1: 49.635}\\
\textbf{ROUGE-2: 25.1852}\\
\textbf{ROUGE-L: 30.6569}
}

\finding{\textbf{\underline{MPT-7B Window Attention (50\% KV Cache)}:}   Chelsea manager Jose Mourinho is known to be an admirer of the midfield player Costa . Chelsea have had Costa watched extensively, while Manchester United and Liverpool have both been linked with a move. Chelsea have had Costa watched extensively, while Manchester United and Liverpool have both been linked with a move. Arsenal have had Costa watched extensively, while Manchester United and Liverpool have both been linked with a move. Chelsea have had Costa watched extensively, while Manchester United and Liverpool have both been linked with a move. Chelsea are set to lead the charge for the 24-year-old having tracked the Brazil international for two years.\\
\textbf{ROUGE-1: 31.5789}\\
\textbf{ROUGE-2: 12.0}\\
\textbf{ROUGE-L: 19.7368}
}

\finding{\textbf{\underline{MPT-7B H$_{2}$O (50\% KV Cache):}}  Douglas Costa will spark a transfer scramble this summer with Shakhtar Donetsk ready to sell their prized asset. Chelsea manager Jose Mourinho is a known admirer of the Brazil international having tried to land the midfielder in the previous transfer windows. Shakhtar chiefs are now open to selling Costa this summer and talks over his departure are underway. Brazil international Douglas Costa will leave Shakhtar Donetsk for £25million in the summer . The 24-year-old midfielder wants to remain playing for a Champions League side and is open to a move to England .\\
\textbf{ROUGE-1: 44.6043}\\
\textbf{ROUGE-2: 24.8175}\\
\textbf{ROUGE-L: 33.0935}
}

\finding{\textbf{\underline{MPT-7B Keyformer (50\% KV Cache:)}}  Douglas Costa will spark a transfer scramble this summer with Shakhtar Donetsk ready to sell their prized-asset. Chelsea manager Jose Mourinho is a known admirer of the Brazil international having tried to land the midfielder in the previous transfer windows. Shakhtar chiefs are now ready to sell Costa this summer and talks over a move to the Premier League are already underway. Brazil international Douglas Costa is set to leave Shakhtar Donetsk this summer with Chelsea, Barcelona and Real Madrid all interested in his signature. The 24-year-old has been linked with a move to the Premier League.\\
\textbf{ROUGE-1: 46.1538}\\
\textbf{ROUGE-2: 28.3688}\\
\textbf{ROUGE-L: 33.5664}
}

\subsection{Extended Related Work}

\paragraph{KV Cache Compression and Quantization:}
Extensive research has explored model compression and quantization to accommodate larger models within limited memory. For autoregressive LLMs, as sequence lengths increase, the size of \kv exceeds the model parameters' size. Implementing compression and quantization techniques on \kv leads to faster inference. FastGen~\cite{ge2024model} introduces adaptive \kv compression by preserving special tokens. ALISA~\cite{zhao2024alisa} utilizes sparse window attention and quantization for \kv. GEAR~\cite{kang2024gear} proposes a \kv compression framework using quantization and low-rank matrices to approximate quantization error. QAQ~\cite{dong2024qaq} introduces quality-adaptive quantization for \kv. DejaVuLib~\cite{strati2024dejavu} implements efficient prompt-token disaggregation to reduce pipeline delays for distributed LLM serving.

\paragraph{Popular Embeddings in Other Large Models:}
The concept of identifying popular or key embeddings (and potentially dropping them) has being investigated for recommendation and other large models~\cite{fae, hotline, fluid}. This concept is used to enable an intelligent embedding placement across heterogeneous memory hierarchies. This helps enhance training efficiency. Similarly, the presence of \key tokens within autoregressive LLMs aids in reducing the \kv size and its allocation within limited GPU memory.

\subsection{Sparsity within Large Language Models}
\label{subsec:sparsity}

\begin{figure}[h!]
  \centering
  \includegraphics[width=0.5\textwidth]{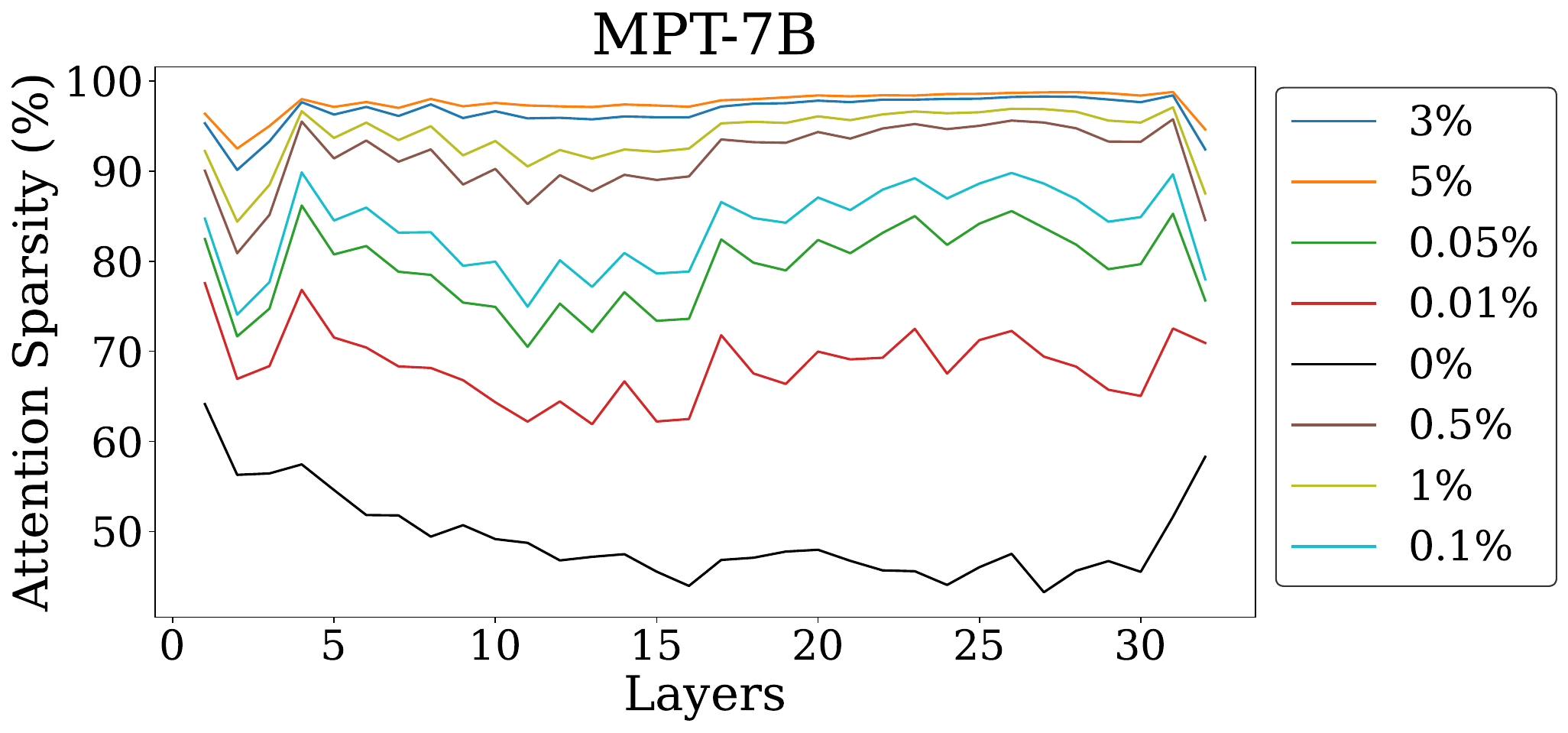}
  \vskip -0.1in
  \caption{Increase in sparsity with varying threshold percentage. 
}
\label{fig:threshold_sparsity}
\end{figure}

We analyzed the MPT-7B model to explore the sparsity of LLMs. Our approach involved examining threshold-based sparsity by varying the percentage threshold of the maximum attention score. We tested thresholds from 0\% to 5\%, where 0\% represents tokens with no attention score. The resulting Figure~\ref{fig:threshold_sparsity} illustrates increased sparsity as the percentage threshold rises.

\subsection{Recent Window versus Key Token Window Ratio}
\label{subsec:recent_vs_key_ratio}

\begin{figure}[h!]
  \centering
  \includegraphics[width=1\columnwidth]{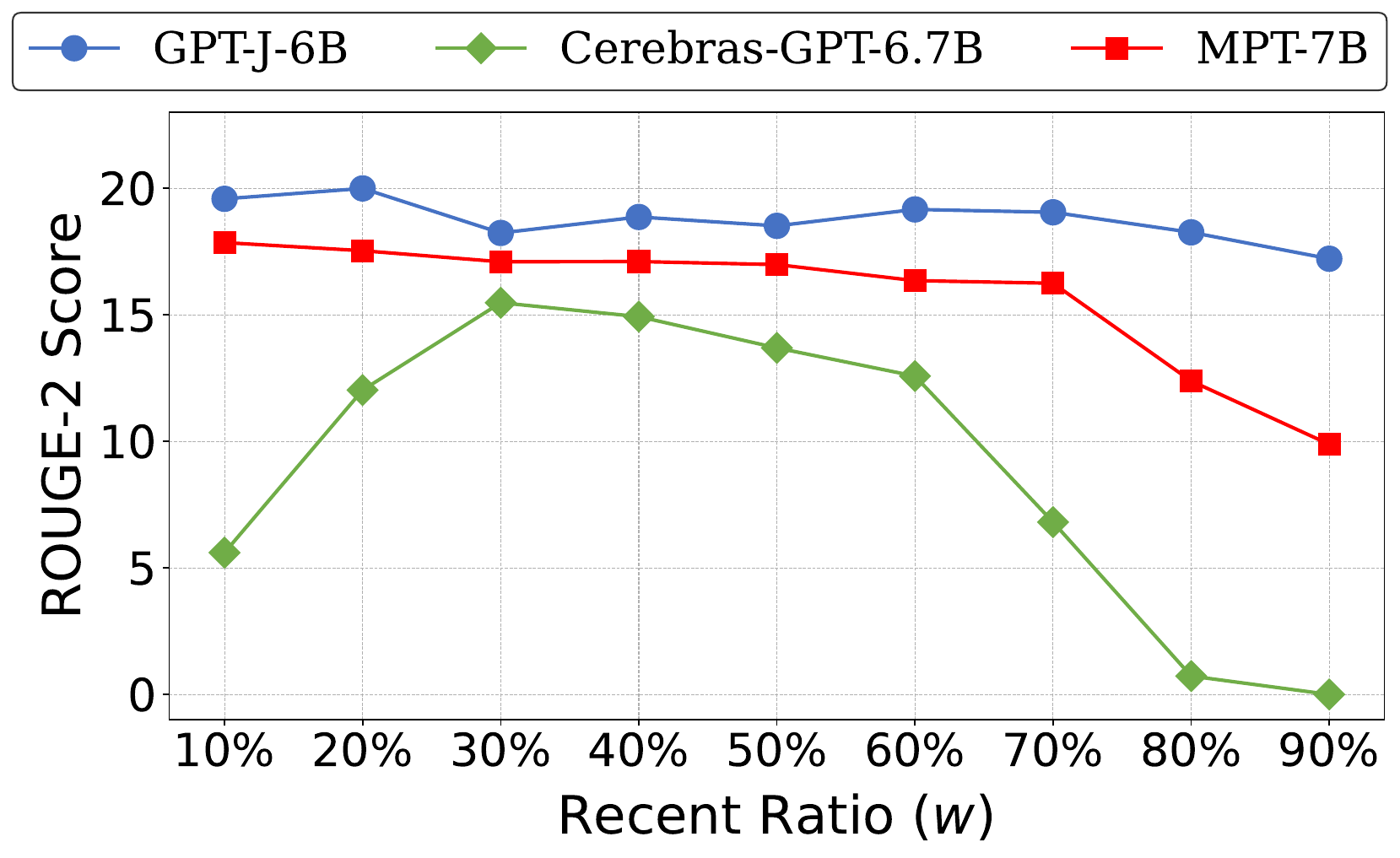}
  \caption{Varying the recent ratio $(w)$ in \keyformer at 70\% \kv and its impact on model accuracy for summarization task with CNN/DailyMail dataset. 
}
\label{fig:recent_ratio}
\end{figure}

We conducted a sensitivity study to examine the impact of varying the ratio of recent tokens $w$ on the size of the \kv dedicated to recent tokens and \key. We maintained a fixed \kv size of 70\%. This resulted in changes in the number of \key $(k-w)$. Figure~\ref{fig:recent_ratio} displays the trend in model accuracy for all three respective models. The results indicate that the models perform better when the recent tokens ratio $w$ falls within the range of 20\% to 30\%. This observation aligns with our hypothesis that both recent and \key tokens tend to be of the highest importance for text generation tasks.

\subsection{ROUGE-1 and ROUGE-L Scores}
\label{subsec:rouge_r}

\begin{figure*}
  \centering
  \includegraphics[width=1\textwidth]{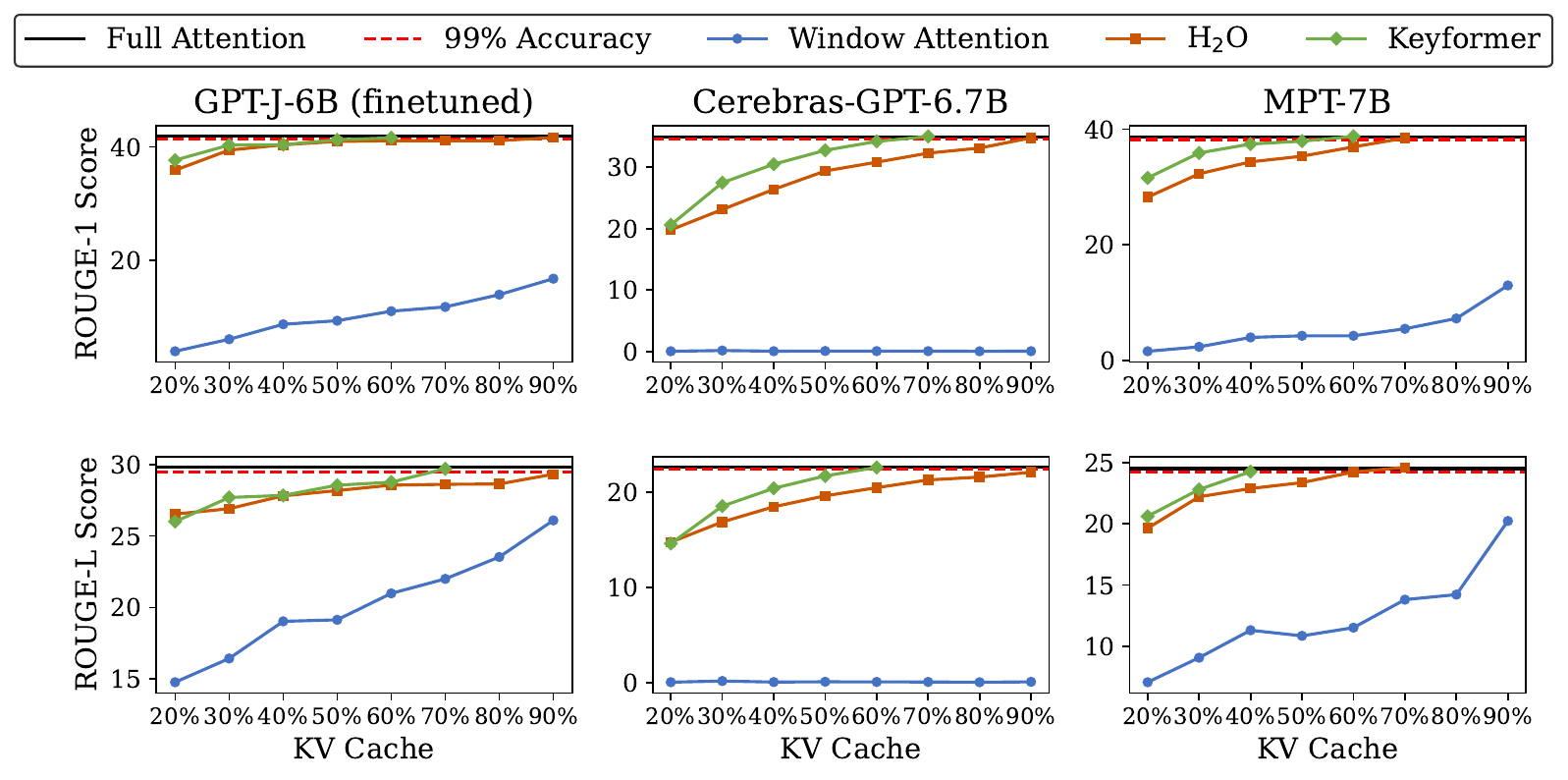}
  \caption{ROUGE-1 and ROUGE-L scores comparison of Full Attention, Window Attention, H$_{2}$O and \keyformer with varying \kv size. The solid black line shows Full Attention without discarding any token and full \kv. The red dotted line shows 99\% accuracy mark as per MLPerf~\cite{mlperf}.
}
\label{fig:accuracy_r}
\end{figure*}

The MLPerf benchmarks~\cite{mlperf} establish rigorous standards for summarization tasks, mandating that all ROUGE scores, encompassing \emph{ROUGE-1, ROUGE-2, and ROUGE-L}, should range between 99\% and 99.9\% of the original scores. Figure~\ref{fig:accuracy_r} shows the trends of ROUGE-1 and ROUGE-L scores in summarization tasks using the CNN/DailyMail validation dataset for three models.

\begin{figure*}
  \centering
  \includegraphics[width=1\textwidth]{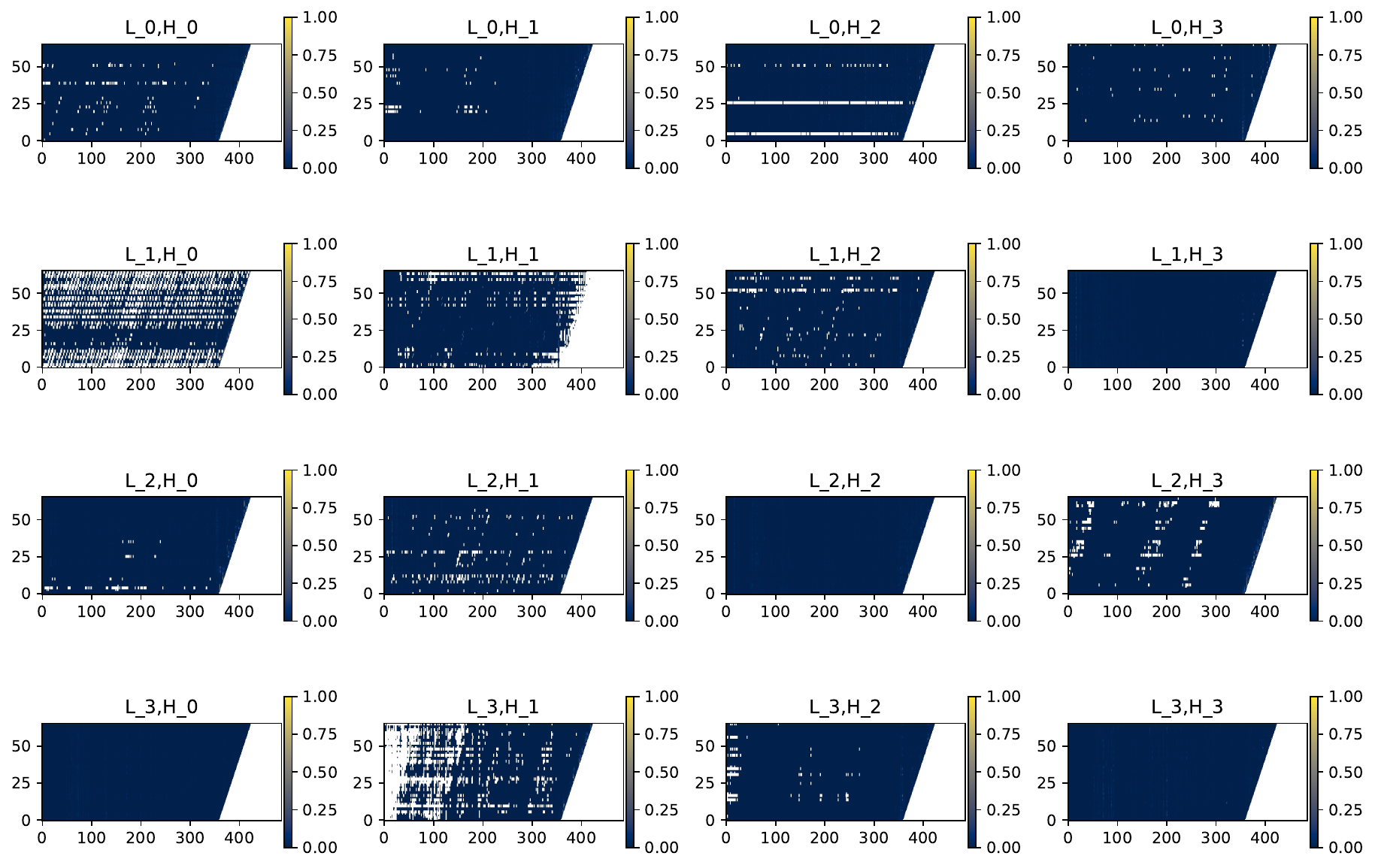}
  \caption{Attention heat map for GPT-J~\cite{gptj} model with first 4 layers and heads (denoted as L\_<layer>,H\_<head>). The x-axis comprises context + text generation, while the y-axis only contains text generation. Empty (white) dots show zero attention score and present inherent sparsity.
}
\label{fig:gptj_attn_map}
\end{figure*}

\begin{figure*}
  \centering
  \includegraphics[width=1\textwidth]{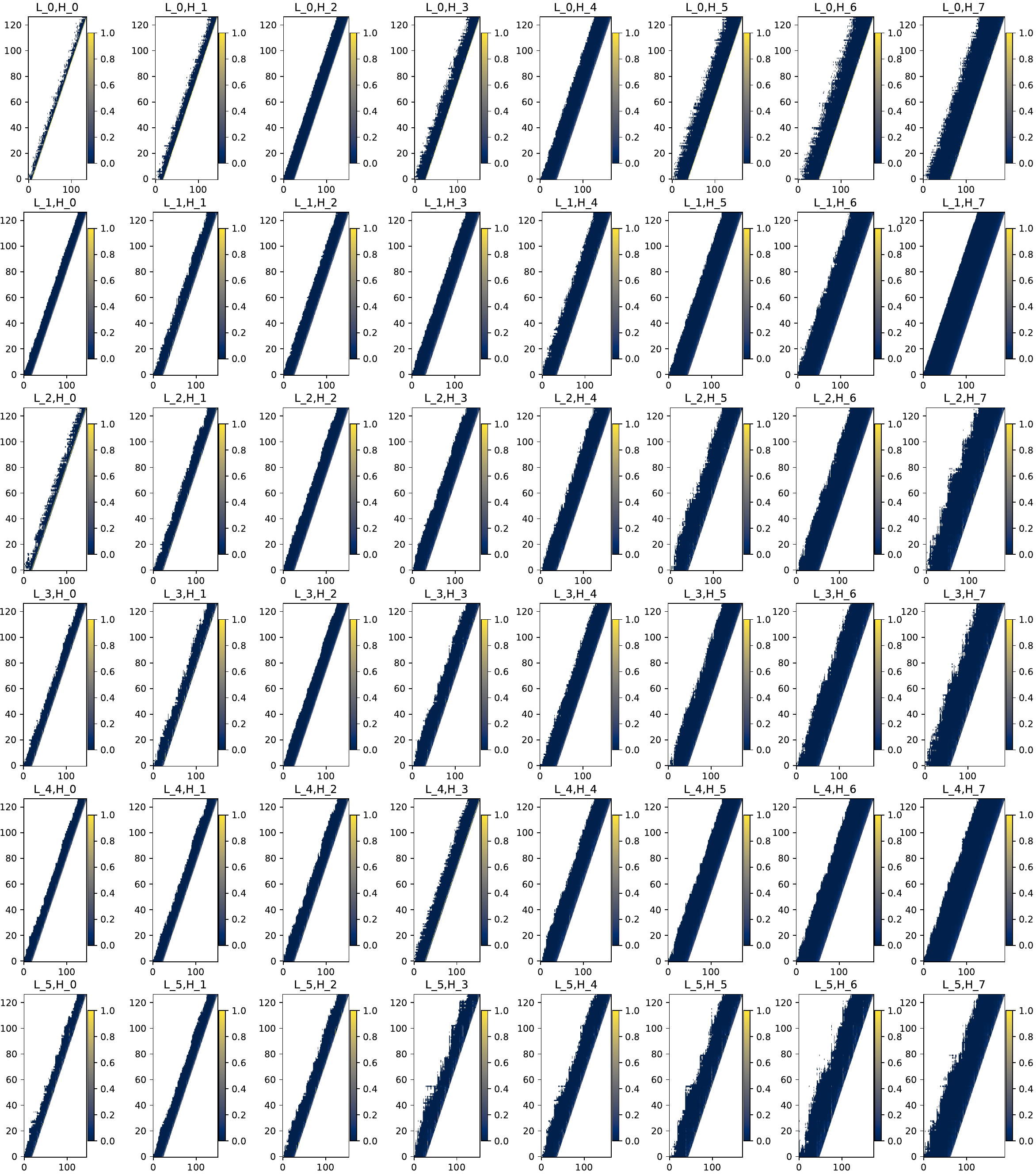}
  \caption{Attention heat map for MPT-7B~\cite{mpt} model with first 6 layers and 8 heads (denoted as L\_<layer>,H\_<head>). The x-axis comprises context + text generation, while the y-axis only contains text generation. Empty (white) dots show zero attention score and present inherent sparsity. The effect of ALiBi~\cite{alibi} can be seen as we move from Head 0 to Head 7.
}
\label{fig:mpt_attn_map}
\end{figure*}

\subsection{Attention Heat Maps of Large Language Models}
\label{subsec:attn_heat_maps}

We examined attention heat maps from different models to explore their inherent sparsity. Figure~\ref{fig:gptj_attn_map} presents the attention heat map for a fine-tuned GPT-J~\cite{gptj} model using a sample from the CNN/DailyMail dataset. The input context comprises 360 tokens, and the model produces a summary of 65 tokens. The heat map reveals that inherent sparsity is dispersed across layers and heads without a distinct pattern. This lack of pattern poses a challenge for exploiting sparsity.

Figure~\ref{fig:gptj_attn_map} and Figure~\ref{fig:mpt_attn_map} depict attention heat maps across different layers and heads for the fine-tuned GPT-J and pre-trained MPT models. The variations observed in the attention heat maps can be attributed mainly to differences in the models' positional encoding schemes.

\subsection{Analyzing Attention Sinks}
\label{subsubsec:attn_sink}

StreamingLLM~\cite{streaming_llm} introduced the concept of ``Attention Sink,'' which prioritizes the initial tokens deemed most important. This approach, which retains the first four tokens alongside the recent window, aims to reduce perplexity effectively. However, as indicated by the attention heat maps presented earlier, \emph{we did not observe a similar trend in our models}.
As explored in ALiBi~\cite{alibi}, the MPT model experimented with training on shorter sequence lengths and testing on longer sequences within the training domain. A constant bias was introduced into the attention mechanism to facilitate the successful generation of longer sequences. This bias gradually diminishes in value as it extends to older generated tokens, with recent tokens receiving a higher bias and tokens further back in the sequence receiving a smaller bias. This bias distribution is observable in the MPT attention heat maps. This characteristic may contribute to the sub-optimal performance of StreamingLLM in summarization tasks.

\subsection{Sensitivity of the Temperature Parameter}
\label{subsubsec:temperature}
\begin{figure}[ht!]
  \centering
  \includegraphics[width=0.9\columnwidth]{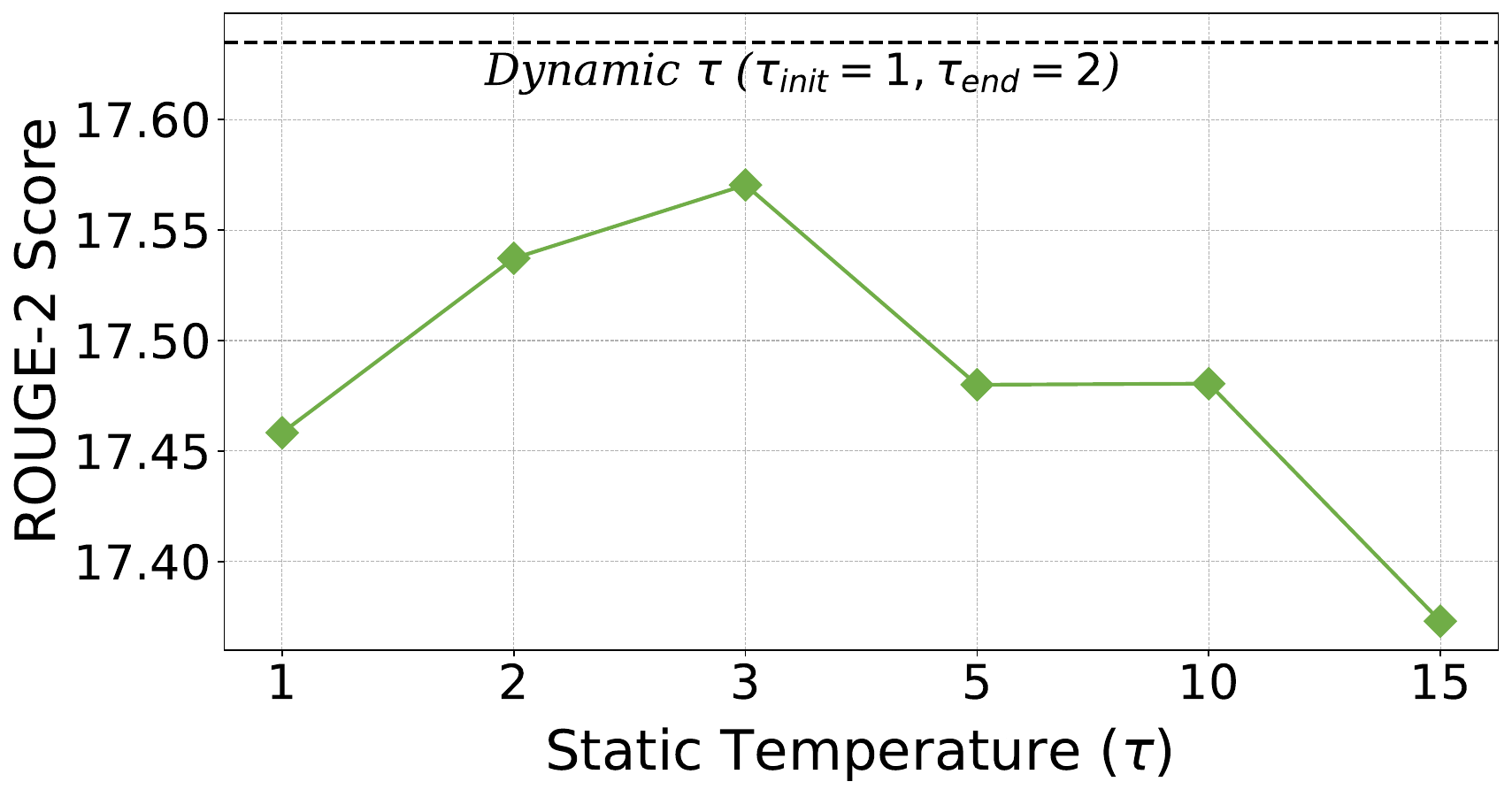}
  \caption{Varying the temperature parameter $(\tau)$ for MPT-7B model for \keyformer and its impact on model accuracy for summarization task with CNN/DailyMail dataset. Sweeping value of $\tau$ from 1 to 2 across the token generation phase works better in comparison to fixed temperature across the prompt processing and token generation phase.
}
\label{fig:temperature}
\end{figure}

We conducted an ablation study to examine the impact of the temperature parameter $\tau$ on \keyformer's performance. This study focused on the summarization task using the MPT-7B model and the CNN/DailyMail dataset. We varied the values of $\tau$ for the Gumbel softmax, maintaining a consistent $\tau$ throughout both the prompt processing and text generation phases. This ensured a uniform level of randomness in the Gumbel softmax distribution, independent of discarded tokens. Moreover, we adjusted the randomness intuitively to match the number of discarded tokens, aiming to compensate for them during text generation iterations. Figure~\ref{fig:temperature} shows how sweeping $\tau$ levels helped identify key tokens and enhance the quality of the model compared to maintaining a fixed $\tau$ value.

\subsection{Artifact Evaluation}

This artifact provides the source code for \keyformer.

\paragraph{Access to Artifact:}

The artifact is available on GitHub at the following link: \href{https://github.com/d-matrix-ai/keyformer-llm}{https://github.com/d-matrix-ai/keyformer-llm}.

\paragraph{Datasets:}

For extensive evaluation across both summarization and conversation tasks, we utilized multiple datasets:

\textbf{CNN/DailyMail}~\footnote{\href{https://huggingface.co/datasets/cnn_dailymail}{https://huggingface.co/datasets/cnn\_dailymail}}: dataset comprises over 300,000 unique news articles written by journalists from CNN and the Daily Mail. It facilitates both extractive and abstractive summarization.

\textbf{GovReport}~\footnote{\href{https://huggingface.co/datasets/ccdv/govreport-summarization?row=0}{https://huggingface.co/datasets/ccdv/govreport-summarization?row=0}}: dataset contains approximately 19.5k U.S. government reports featuring expert-written abstractive summaries. It includes notably longer documents (9.4k words) and summaries (553 words) compared to other existing datasets.

\textbf{SODA}~\footnote{\href{https://huggingface.co/datasets/allenai/soda}{https://huggingface.co/datasets/allenai/soda}}: dataset is a high-quality collection of dialogues encompassing diverse social interactions.

The validation set of each dataset is utilized for evaluation.

\paragraph{System Requirements:}

For evaluation, the following hardware is employed:

\begin{itemize}
    \item 180 GB of disk space
    \item 90 GB of DRAM
    \item NVIDIA Tesla A100 (80~GB) GPU
\end{itemize}